\documentclass{article}

\usepackage{microtype}
\usepackage{graphicx}
\usepackage{subcaption}
\usepackage{booktabs} 

\usepackage{custom}

\usepackage{hyperref}

\newcommand{\para}[1]{\smallskip\noindent{\textbf{#1}}}

\usepackage[preprint]{icml2026}

\usepackage{amsmath}
\usepackage{amssymb}
\usepackage{mathtools}
\usepackage{amsthm}

\usepackage[capitalize,noabbrev]{cleveref}

\theoremstyle{plain}

\theoremstyle{definition}

\theoremstyle{remark}

\usepackage[textsize=tiny]{todonotes}

\icmltitlerunning{\sys: Generative Code Representation Learning with Code Transformations}

\begin{document}\sloppy

\newcommand{\kexin}[1]{{\color{red}{(\framebox{Kexin:} #1)}}}
\newcommand{\weichen}[1]{{\color{blue}{(\framebox{Weichen:} #1)}}}
\newcommand{\jiamin}[1]{{\color{orange}{(\framebox{Jiamin:} #1)}}}
\newcommand{\bo}[1]{{\color{violet}{(\framebox{Bogdan:} #1)}}}
\newcommand{\gabe}[1]{{\color{brown}{(\framebox{Gabe:} #1)}}}

\newcommand{\sys}{\textsc{SemRep}\xspace}
\newcommand{\sem}{\textsc{sem}\xspace}
\newcommand{\proto}{\hat{\mathbf{x}}\xspace}
\newcommand{\obs}{\mathbf{x}\xspace}

\twocolumn[
  \icmltitle{\sys: Generative Code Representation Learning with Code Transformations}



  \icmlsetsymbol{equal}{*}

  \begin{icmlauthorlist}
    \icmlauthor{Weichen Li}{uchicago}
    \icmlauthor{Jiamin Song}{uchicago}
    \icmlauthor{Bogdan Alexandru Stoica}{uiuc}
    \icmlauthor{Arav Dhoot}{columbia}
    \icmlauthor{Gabriel Ryan}{msft}
    \icmlauthor{Shengyu Fu}{msft}
    \icmlauthor{Kexin Pei}{uchicago}
  \end{icmlauthorlist}

  \icmlaffiliation{uchicago}{The University of Chicago}
  \icmlaffiliation{uiuc}{University of Illinois Urbana-Champaign}
  \icmlaffiliation{columbia}{Columbia University}
  \icmlaffiliation{msft}{Microsoft}

  \icmlcorrespondingauthor{Weichen Li}{weichenli@uchicago.edu}
  \icmlcorrespondingauthor{Kexin Pei}{kpei@uchicago.edu}

  \icmlkeywords{Machine Learning, ICML}

  \vskip 0.3in
]



\printAffiliationsAndNotice{}  

\begin{abstract}
    Code transformation is a foundational capability in the software development process, where its effectiveness relies on constructing a high-quality code representation to characterize the input code semantics and guide the transformation.
    Existing approaches treat code transformation as an end-to-end learning task, leaving the construction of the representation needed for semantic reasoning implicit in model weights or relying on rigid compiler-level abstractions.
    We present \sys, a framework that improves code transformation through \emph{generative code representation learning}.
    Our key insight is to employ the semantics-preserving transformations as the intermediate representation, which serves as both a generative mid-training task and the guidance for subsequent instruction-specific code transformations.
    Across general code editing and optimization tasks (e.g., GPU kernel optimization), \sys outperforms the extensively finetuned baselines with strictly the same training budget by 6.9\% in correctness, 1.1$\times$ in performance, 13.9\% in generalization, and 6.7\% in robustness.
    With the improved exploration of diverse code transformations, \sys is particularly amenable to evolutionary search.
    Combined with an evolutionary coding agent, \sys finds optimizations that 685B larger-weight baselines fail to discover while achieving the same performance with 25\% less inference compute.
    
\end{abstract}

\section{Introduction}

Code transformation is a foundational step in the software engineering process, enabling a wide variety of software development and maintenance workflows.
To automate code transformation, Large Language Models (LLMs) have been increasingly adopted to assist developers in concrete code editing tasks, such as efficiency optimization~\cite{shetty2025gso, ma2025swe, he2025swe}, refactoring and migration~\cite{ziftci2025migrating, eniser2024towards}, bug fixing~\cite{jimenez2023swebench, yu2025patchagent}, and feature integration~\cite{chi2025editbenchevaluatingllmabilities, cassano2023can, nam2025prompting}.
Emerging evidence also suggests that LLM-driven code transformation can discover high-performance GPU kernels~\cite{ouyang2025kernelbench, wei2025astra}, co-design system architectures~\cite{cheng2025let, cheng2025barbarians}, and even serve as a substrate for algorithmic and scientific discovery~\cite{press2025algotune, novikov2025alphaevolve, lange2025shinkaevolve}.

\begin{figure}[!t]
    \centering
    \includegraphics[width=1\linewidth]{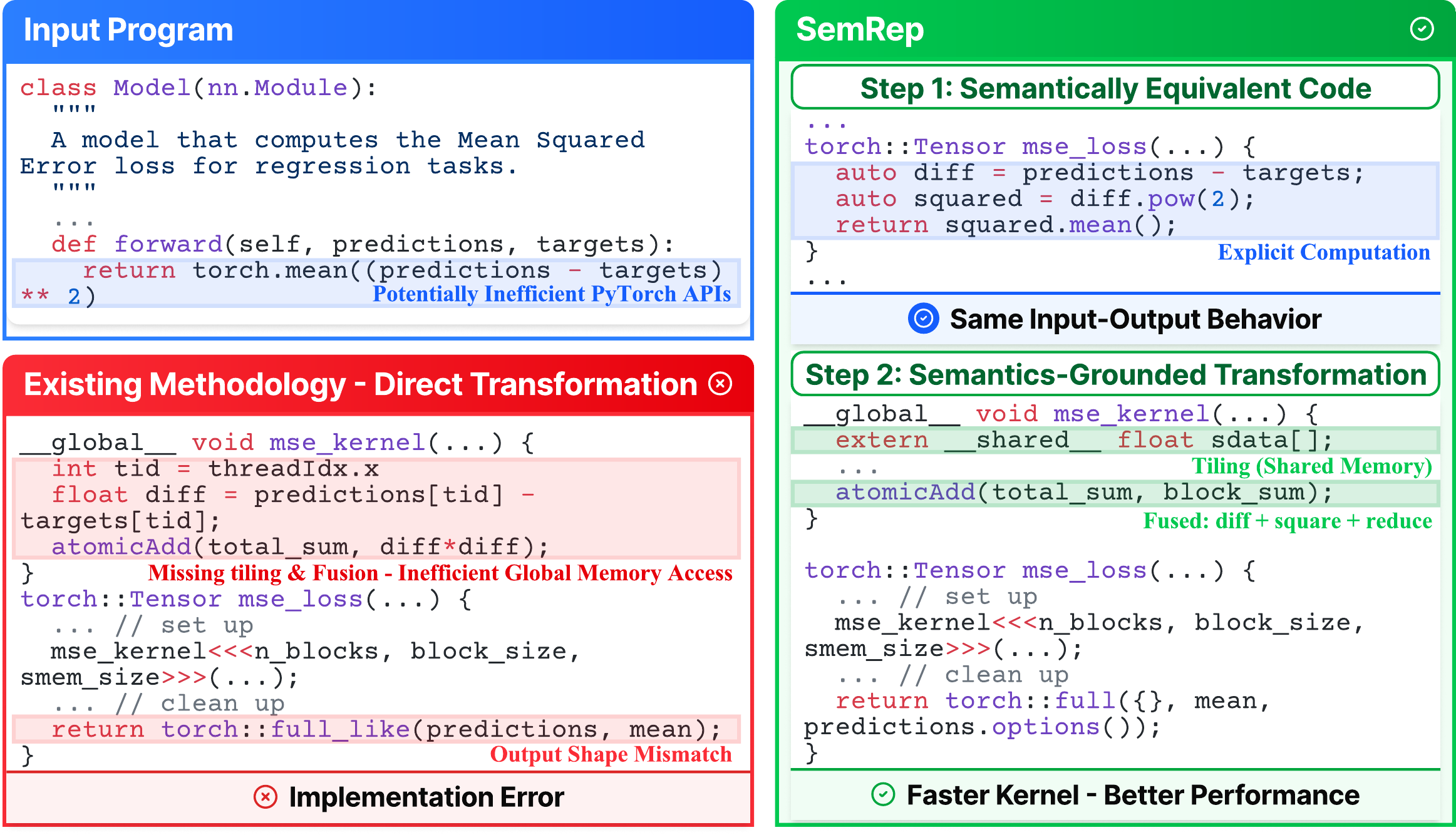}
    \caption{A concrete example showing how \sys optimizes CUDA kernels in Kernelbench~\cite{ouyang2025kernelbench}. 
    \sys first generates a semantically equivalent code that explicitly exposes the computation of MSE loss, and then transforms it into a more efficient implementation that optimizes memory access by replacing separate passes over large arrays with a single pass, where each thread computes its global index, reads inputs into fast registers for local computation, and fuses multiple operators into a single kernel execution. The existing approach attempts to directly optimize MSE but fails on several core operations.}
    \label{fig:motivation}
\end{figure}

A key capability for effective code transformation is to disentangle \textit{what must be preserved} from \textit{what can be changed}.
This often requires the model to understand the intended behavior of the original code, i.e., its semantics, and to reason about the effects introduced by the transformation.
\cref{fig:motivation} illustrates why this matters: without recognizing that the input code performs a matrix reduction, state-of-the-art approaches~\cite{baronio2025kevin} generate incorrect edits while missing the optimization opportunity to fuse operations and optimize memory access patterns.

To characterize the intended behavior of the original code and support desired transformations, automated tools often explore intermediate code \emph{representations} to expose the code properties, such as those used in compilers like abstract syntax trees, 
or control/data flow graphs, where a fixed set of compiler transformations on top of these representations is well defined and can be efficiently implemented.

Unfortunately, when extending the support of compiler transformations to broader natural language-instructed code editing, existing works often degenerate to be entirely data-driven, treating the transformation as an end-to-end learning task while relying on the model to \emph{implicitly learn representations necessary to code transformation as latent weight parameters}~\cite{pie_iclr_2024_spotlight, tan2024llm4decompile, wei2025swe}.
While some works explicitly incorporate more structured representations in the model~\cite{allamanis2017learning, pei2023exploiting}, constructing such representations on the fly can be expensive.
More importantly, it is unclear whether these rigid representations remain \emph{informative} for code LLMs (which are predominantly pre-trained on source code rather than compiler-level abstractions), and \emph{optimal}, as a single structural representation suitable for transforming an input code may not necessarily generalize to another.

\begin{figure*}[!t]
    \centering
    \includegraphics[width=1.0\linewidth]{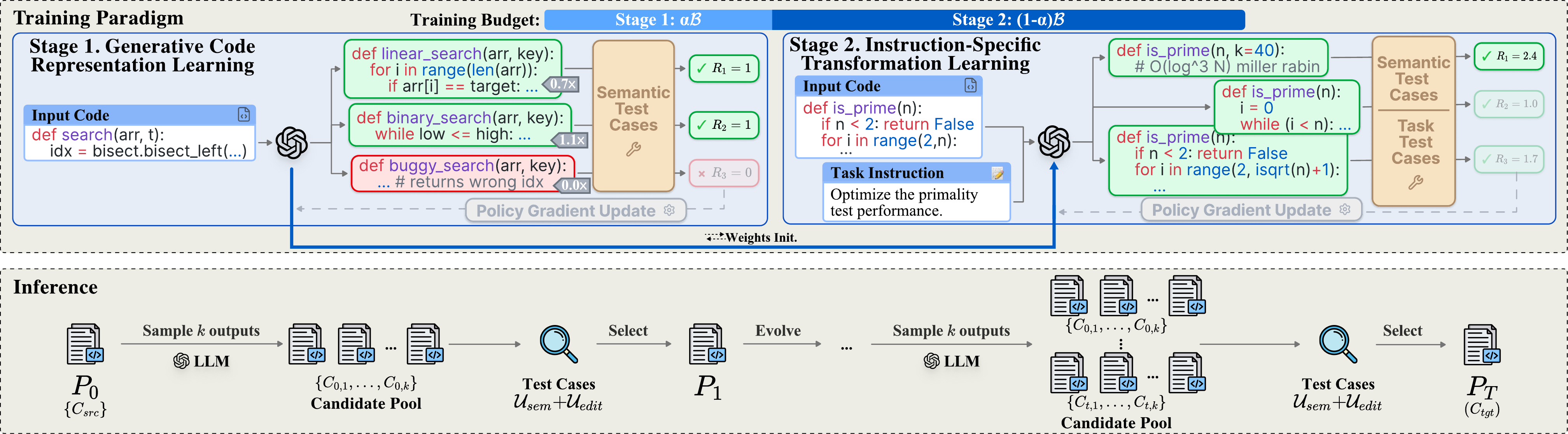}
    \caption{\sys framework. 
    The upper shows how \sys trains a model to instill semantics understanding through the generative code representation learning. 
    \sys explicitly encourages diverse semantically equivalent implementations, e.g., both faster and slower code variants, even when editing objectives are not fully satisfied.
    The total training budget $\mathcal{B}$ is fixed to ensure the fair comparison to those finetuning-only baselines. 
    $\alpha$ modulates the training budget allocated to Stage 1 and Stage 2.
    The lower figure shows how \sys performs iterative inference.}
    \label{fig:framework}
\end{figure*}

\para{Our approach.}
We present \sys, a new framework that improves code transformation capabilities by training LLMs to explicitly learn code representations as semantics-preserving transformations.
Specifically, we define a mid-training task, \emph{generative code representation learning}, by training the model to generate semantically equivalent programs, i.e., producing the same execution outputs as the original program for given inputs~\cite{le2014compiler}.
During inference, given the input code, the trained model is allowed to alternate between the generation of (1) semantically equivalent code snippets as part of the \emph{exploration}, and (2) transformed code following the initial instructions, based on the code generated in (1).
Figure~\ref{fig:motivation} shows how \sys generates a single semantically equivalent code, which inspires the generation of the optimized CUDA kernel.
Figure~\ref{fig:framework} shows its overall workflow.

\sys's task formulation encourages the model to produce explicit code representations, i.e., semantically equivalent source code, that remain directly \emph{interpretable} by LLMs and humans, but also amenable to \emph{unambiguous} code reasoning tools, such as execution~\cite{ding_semcoder_2024, copet2025cwm} and static analysis~\cite{van_tonder_tailoring_2020, chen2018learning, namjoshi2018impact}.
The former enables the generation of \emph{diverse and human-interpretable structured reasoning}, i.e., chain of code~\cite{li2023chain}, to enable the generation of (potentially more optimal) subsequent instruction-specific code transformation, while the latter ensures such a structured reasoning is \emph{verifiably correct} (modulo test input~\cite{le2014compiler}).

Such a design offers several benefits when guiding LLMs' training and inference for code transformation.
First, it teaches the model to discover representations that express code semantics \emph{explicitly}, as opposed to latent representation in model weights.
Second, it exposes opportunities for \emph{diverse} and \emph{exploratory} code transformations, which are particularly amenable for test-time scaling like evolutionary search, where exploration is the key to finding an optimal solution~\cite{novikov2025alphaevolve}.
Third, the verifiable rewards are particularly well-suited to reinforcement learning, sharing a similar spirit to reinforcement pre-training~\cite{dong_reinforcement_2025, hatamizadeh_rlp_2025} that is \emph{easy to scale}.
Lastly, the generative formulation of semantically equivalent code often aligns with most of the practical code editing setups, where it is required to preserve the original code functionality while introducing new properties, such as performance, readability, and security~\cite{li2025editlord}.

We implement our representation mid-training using reinforcement learning (RL) with verifiable rewards~\cite{guo2025deepseek, shao2024deepseekmath} based on test execution, and then finetune (via RL) the model for various code transformation tasks like CUDA kernel optimizations~\cite{ouyang2025kernelbench} and general-purpose code editing~\cite{chi2025editbenchevaluatingllmabilities}.
To show the unique advantage of the generative representation learning, we construct a dedicated baseline that adopts the same model as \sys but only finetuned for the code transformation (not mid-trained) with the \emph{exactly same training budget}, i.e., the total iterations of mid-training and finetuning in \sys matches this baseline's finetuning iterations.
This ensures that we do not introduce extra compute when training for representation learning.

\para{Results.}
Our evaluation shows that \sys enables smaller models, e.g., QwQ-32B~\cite{qwq32b}, to match or outperform models with 12$\times$ larger weights, and closed-weight commercial LLMs, across all tasks and inference paradigms.
\sys outperforms the extensively finetuned baselines with strictly the same training budget by 6.9\% in correctness and 1.1$\times$ in performance.
With explicit representation learning, \sys significantly improves the baseline in generalization, e.g., by up to 13.9\% in cross-hardware kernel optimization, and 6.7\% improved robustness against semantics-preserving code transformations. 
When integrating \sys with advanced evolutionary coding agent~\cite{openevolve}, it is able to optimize state-of-the-art algorithms that strong baselines (e.g., DeepSeek-V3-Reasoner) fail to find, while achieving the same performance with 25\% less compute.

\section{Methodology}
\label{sec:method}

\newcommand{\src}{\text{$C_{src}$}\xspace}
\newcommand{\tgt}{\text{$C_{tgt}$}\xspace}
\newcommand{\rep}{\text{$C_{rep}$}\xspace}

\para{Problem statement.}
We formulate the code editing task as the transformation of a source program \src into a target program \tgt conditioned on a natural language instruction $I$. 
This transformation must meet two distinct requirements: (1) semantic preservation of existing logic, and (2) alignment with the new instruction.

We define the semantics-preserving transformation using test-based equivalence~\cite{le2014compiler}.
Given a set of tests $\mathcal{U}$, two programs $C_1$ and $C_2$ are considered equivalent modulo the tests $\mathcal{U}$, denoted as 
$ C_1(\mathcal{U}) =  C_2(\mathcal{U})$, if they produce identical outputs for all tests in $\mathcal{U}$.

We use two test sets, $\mathcal{U}_{sem}$ and $\mathcal{U}_{edit}$, to measure semantic preservation and instruction adherence, respectively.
The goal is to learn a conditional distribution $P(\tgt| \src, I)$, subject to 
$\src(\mathcal{U}_{sem})=\tgt(\mathcal{U}_{sem})$, 
and \tgt satisfies the requirements specified in $I$ that is verifiable by $\mathcal{U}_{edit}$.

The challenge in code editing is to disentangle what needs to be preserved (e.g., the original functionality of \src) with respect to $\mathcal{U}_{sem}$, and what needs to be edited (the new requirement specified in $I$) with respect to $\mathcal{U}_{edit}$.

To address this challenge, \sys decomposes the editing process into the transition:
$\src\rightarrow\dots\rightarrow\rep\rightarrow\dots\rightarrow\tgt$, 
where \rep serves as an explicit and verifiable intermediate semantic representation that satisfies 
$\rep(\mathcal{U}_{sem})=\src(\mathcal{U}_{sem})$, before further guided changes are applied.
These transitions are learned in two stages:
(1) Generative code representation learning trains the model to generate semantically equivalent program variants.
(2) Instruction-specific transformation finetuning trains the model to transform programs to align with $I$.

\subsection{Generative Code Representation Learning}
\label{subsec:method_mid_training}
In the first stage of \sys's training, we optimize a base model 
to generate semantics-preserving variants of \src.
The goal is to produce a representation \rep that explores the semantic equivalent class of \src that stays functionally invariant under the tests $\mathcal{U}_{sem}$.
We formulate this as a reinforcement learning task using Group Relative Policy Optimization (GRPO)~\cite{shao2024deepseekmath},
and define the reward $R_{sem}$ that measures semantic invariance while encouraging the model to explore different syntax or structures: 
\begin{align*}
R_{sem}(\rep) &= \alpha_1 \cdot \mathbb{I}_{comp}[\rep] \\
&\quad +\beta_1\cdot\mathbb{I}_{\mathcal{U}_{sem}}\big[\rep=\src\big]
\end{align*}
Here $\alpha_1$ and $\beta_1$ are weights that modulate the focus between compilability (as it is a pre-requisite for the code to be executable using $\mathcal{U}_{sem}$) and semantic invariance.
$\mathbb{I}_{comp}(\rep)$ denotes successful compilation, and 
$\mathbb{I}_{\mathcal{U}_{sem}}[\rep = \src]$ 
is the indicator function for equivalence based on the test set $\mathcal{U}_{sem}$. 
This reward design is intended to encourage exploring diverse, semantically equivalent variants of the given source code during training (e.g., different loop structures or API calls) that elicit creative transformation possibilities in the downstream editing task, e.g., code optimizations.

\subsection{Instruction-Specific Transformation Learning}
\label{subsec:method_finetuning}
In the second stage of \sys's training, we use an instruction-specific editing objective, where the model must introduce the changes requested by $I$ while preserving the original program's intended behavior unaffected by the edit.

Similar to~\cref{subsec:method_mid_training}, we employ GRPO and optimize for an instruction-specific reward $R_{inst}$ that jointly rewards instruction-following and semantics-preserving changes in the transformed program:  
\begin{align*}
R_{inst}(\tgt) &= \alpha_2 \cdot \mathbb{I}_{comp}[\tgt] \\
& + \beta_2 \cdot \quad \mathbb{I}_{\mathcal{U}_{sem}}\big[ \src = \tgt \big] \\
&\quad + \gamma \cdot \mathbb{I}_{\mathcal{U}_{edit}}\big[C_{tgt}\big]
\end{align*}
Here $\alpha_2,\beta_2,\gamma$ control the compilability, semantic invariance, and instruction adherence. $\mathbb{I}_{comp}[\tgt]$ denotes successful compilation, 
$\mathbb{I}_{\mathcal{U}_{sem}}[\src = \tgt]$ 
measures equivalence, and $\mathbb{I}_{\mathcal{U}_{edit}}[C_{tgt}]$ verifies instruction adherence.

\subsection{Iterative Inference and Test-Time Scaling}
We scale the inference by an iterative evolutionary search,
where the model alternates between semantic exploration, i.e., generating diverse semantically equivalent code, and instruction-specific transformation, i.e., optimization, feature integration, etc.
This generates a sequence of program pools $P_0, P_1, \dots, P_T$, where $P_0 = \{\src\}$, and $\forall t\in[1, T], P_t=\{c_1, \cdots, c_{b}\}$ maintains a beam that contains $b$ programs, i.e., number of programs retained at the end of each iteration.
During each turn $t\in[1, T]$, we sample $k$ transformed candidates for each program in $P_{t-1}$ by querying the model, resulting in $k
\cdot b$ candidate programs: $\tilde{P}_t = \{\tilde{c}_{1}, \dots, \tilde{c}_{k\cdot b}\}$. 
From $\tilde{P}_t$, we keep only top-$b$ selected programs based on their execution against $\mathcal{U}_{sem}$ and $\mathcal{U}_{edit}$:
$$P_{t} = \underset{c \in \tilde{P_t}}{\text{Top-}b} \big[ \omega_1 {\mathcal{U}_{sem}}[c =\src] + \omega_2 \mathcal{U}_{edit}(c) \big]$$
Here $\omega_1$ and $\omega_2$ are weights that modulate the focus between semantic exploration (finding functionally equivalent variants) and instruction-following editing (realizing editing requests).
Importantly, the capability of generating semantic-preserving programs (not necessarily optimized or fully aligned with user instructions) as intermediate steps facilitates the discovery of more source programs that could potentially lead to more effective edits in downstream tasks.

The final target program $\tgt$ is sampled from $k$ trajectories maintained across $T$ iterations $P_1, \dots, P_T$, and selected based on their execution against $\mathcal{U}_{sem}$ and $\mathcal{U}_{edit}$, e.g., ranked in speedup for optimizations with functionality guarantee.

\section{Evaluation}
\label{sec:evaluation}
We evaluate \sys on two critical software engineering applications that can be formulated as code editing tasks (\cref{subsec:eval_setup}). 
We consider inference using the finetuned model as our default editing mode and compare it to the state-of-the-art baselines, which are also mostly based on finetuned models~\cite{baronio2025kevin}.
We choose models that can be full-parameter RL-finetuned with FSDP~\cite{zhao2023pytorch} on our local hardware (2x4 Nvidia L40S GPUs), i.e., Qwen2.5-Coder-7B~\cite{hui2024qwen2}, and QwQ-32B~\cite{qwq32b}.
We also include multiple case studies on KernelBench (Appendix \ref{app:kernelbench_case_study}), Editbench (\cref{subsec:case_study}), integration with evolutionary coding agent (\cref{subsec:eval_evolve_agent}), and repo-level analysis (Appendix \ref{app:sweperf_case_study}).
\begin{table}[!t]
    \centering
    \small
    \setlength{\tabcolsep}{5pt}
    \renewcommand{\arraystretch}{1}
    \small
    \caption{Resolve rate (pass@1, \%) of \sys and other representative models on EditBench~\cite{chi2025editbenchevaluatingllmabilities} leaderboard using Core set. \sys uses QwQ-32B as a base model, and we report the single run for comparability.}
    \label{tab:main_result_editbench}
    \begin{tabular}{r|rl}
        \toprule
        Model & Model Size & Pass@1 $\uparrow$ \\
        \midrule
        \multicolumn{3}{l}{\textit{Closed Weight Models}} \\
        \midrule
        Claude Sonnet 4                &    -   &     66.67 \\
        GPT o4-mini                    &    -   &     57.41 \\
        Gemini 2.5 Pro                 &    -   &     54.63 \\
        \midrule
        \multicolumn{3}{l}{\textit{Open Weight Models}} \\
        \midrule
        \rowcolor{purple!8} \sys QwQ-32B (Ours)  & 32B & \textbf{57.41} \\
        Qwen3-Coder                    &    405B   &     55.56 \\
        GLM-4.6                        &    355B   &     55.56 \\
        Finetuned QwQ-32B              &    32B &     53.70 \\
        Qwen2.5-72B-Instruct           &    72B   &     53.70 \\
        QwQ-32B                        &    32B &     50.93 \\ \midrule
        \rowcolor{purple!8} \sys Qwen2.5-Coder-7B   & 7B &  32.41\\
        gemma-3-12b-it                 &    12B & 23.15 \\
        Finetuned Qwen2.5-Coder-7B  & 7B & 23.15\\
        Qwen2.5-Coder-7B    & 7B & 18.52\\
        \bottomrule
    \end{tabular}
\end{table}

\begin{table*}[!t]    
    \centering
    \setlength{\tabcolsep}{5pt}
    \renewcommand{\arraystretch}{1}
    \small
    
    \caption{Main results for domain-specific code editing on KernelBench.}
    \label{tab:main_result_kernelbench}
    \begin{tabular}{rllllllll}
    \toprule
    \multirow{2}{*}{} 
        & \multicolumn{2}{c}{Correctness$\uparrow$} 
        & \multicolumn{2}{c}{Speedup$\uparrow$} 
        & \multicolumn{2}{c}{fast$_1^{\ast}\uparrow$}
        & \multicolumn{2}{c}{fast$_{1.5}^{\ast}\uparrow$}  \\
    \cmidrule{2-9}
     & best@16 & avg@16 & best@16 & avg@16 
     & best@16 & avg@16 & best@16 & avg@16\\
     \midrule
     QwQ-32B & 33 & 4.13 & 1.25 & 1.02 & 14 & 2.06 & 4 & 0.69\\
     GPT 4o-mini
        & 47$_{+14}$ & 17.13$_{+13}$ 
        & 1.27$_{+0.02}$ & 1.12$_{+0.1}$ 
        & 22$_{+8}$ & 7.75$_{+5.69}$ 
        & 8$_{+4}$ & 2.13$_{+1.44}$ \\
     Kevin-32B 
        & 65$_{+32}$ & 19.63$_{+15.5}$ 
        & 1.36$_{+0.11}$ & 1.1$_{+0.08}$ 
        & 21$_{+7}$ & 25$_{+22.94}$ 
        & 9$_{+5}$ & 2.69$_{+2}$\\
     \rowcolor{purple!8} 
     \textbf{\sys (Ours)}
        & \textbf{93$_{+60}$} & \textbf{37.63$_{+33.5}$} 
        & \textbf{2.87$_{+1.62}$} & \textbf{1.21$_{+0.19}$} 
        & \textbf{72$_{+58}$} & \textbf{21.06$_{+19}$} 
        & \textbf{12$_{+8}$} & \textbf{3.13$_{+2.44}$}\\  
    \bottomrule
    \end{tabular}

    \vspace{2pt}
    \footnotesize{$^{\ast}$fast$_1$ / fast$_{1.5}$: percentage of trajectories containing at least one correct kernel with speedup $>1$ / $>1.5$, respectively.}

    \setlength{\tabcolsep}{3.5pt}
    \renewcommand{\arraystretch}{1.1}
\end{table*}

\begin{table*}[!t]    
    \centering
    \setlength{\tabcolsep}{5pt}
    \renewcommand{\arraystretch}{1}
    \small
    
    \caption{Ablations on KernelBench and EditBench.}
    
    \label{tab:ablation_study}
    \begin{tabular}{r|llllllll|ll}
    \toprule
    &\multicolumn{8}{c}{KernelBench} &\multicolumn{2}{c}{EditBench}  \\
    \cmidrule{2-11}
    \multirow{2}{*}{}&\multicolumn{2}{c}{Correctness$\uparrow$} &\multicolumn{2}{c}{Speedup$\uparrow$} &\multicolumn{2}{c}{fast$_1\uparrow$}&\multicolumn{2}{c}{fast$_{1.5}\uparrow$} & \multirow{2}{*}{Pass@1} & \multirow{2}{*}{Pass@16}\\
      & best@16 & avg@16 & best@16 & avg@16 & best@16 & avg@16 & best@16 & avg@16 \\
     \midrule
     \textit{Baseline} & 74 & 12.56 & 1.33 & 1 & 45 & 6.75 & 8 & 0.94 & 53.70 & 55.56\\
     \textit{+ \sys TTS} & 59 & 15.19 & 1.47 & 1.13 & 28 & 8.19 & 11 & 3.62 & 56.48 & 62.04\\
     \textit{+ \sys-trained} & 79 & 25.75 & 1.48 & \textbf{1.25} & 35 & 13.5 & 11 & \textbf{5.75} & 55.56 & 62.96\\
     \rowcolor{purple!8} \textbf{\textit{+ Both} (Ours)} & \textbf{93} & \textbf{37.63} & \textbf{2.87} & 1.21 & \textbf{72} & \textbf{21.06} & \textbf{12} & 3.13 & \textbf{57.41} & \textbf{66.67}\\  
    \bottomrule
    \end{tabular}
\end{table*}

\subsection{Setup: Tasks, Datasets and Metrics}
\label{subsec:eval_setup}
\para{Tasks and benchmarks.} 
We evaluate \sys on EditBench~\cite{chi2025editbenchevaluatingllmabilities} and KernelBench~\cite{ouyang2025kernelbench}.
EditBench is constructed from real-world repo-level edits to evaluate how well LLMs follow developer instructions (in natural language) to perform diverse software maintenance operations, e.g., bug fixes, feature modifications or enhancements, and code refactoring.
KernelBench focuses on optimizing PyTorch implementations with customized low-level CUDA kernels to improve performance.

\para{Metrics.}
For EditBench, we report performance using the primary metric established in the leaderboard: 
\emph{Pass@k}: The percentage of problems where one of the model's $k$ generated edits passes all associated unit tests. 
This provides a rigorous assessment of functional correctness, evaluating the model’s ability to implement intended changes and produce deployable code on the first attempt.

For KernelBench, we adopt the same metrics as Kevin~\citep{baronio2025kevin}:
\emph{Correctness}: whether an LM-generated trajectory contains at least one correct kernel.
\emph{Speedup}: the absolute ratio between the execution time required by the given PyTorch code and the fastest and correct code within an LM-generated trajectory.
\emph{fast$_p$}: whether there exists a correct code that achieves speedup $>p$ within the given trajectory.
For each metric, we report two aggregations across $k$ parallel trajectories:
\emph{Best@k}: The maximum value achieved across $k$ trajectories.
\emph{Avg@k}: The arithmetic mean of the metric across all $k$ trajectories.

\para{Training and inference.}
We train \sys with verl~\cite{sheng2024hybridflow} and use GRPO~\cite{shao2024deepseekmath} without a KL penalty following \citet{baronio2025kevin, hatamizadeh_rlp_2025}. 
To ensure a fair comparison, we enforce a \emph{strictly same} training budget across all experiments, including baseline finetuning and our proposed \sys, to prevent overfitting (detailed in Appendix~\ref{app:budget}).

Due to computational budget constraints, we limit our inference to $T=2$, $b=1$, and $k=16$ parallel rollouts at each iteration, though arbitrary $T,k$ and $b$ are supported.
For the specific case of $T=2$, the first iteration explores semantically equivalent variants \rep based on the given \src, while the second iteration applies instruction-specific transformation to generate \tgt from the selected \rep based on $\mathcal{U}_{sem}$.

\subsection{Main Results}
\label{subsec:eval_main_results}
We compare \sys to original models and state-of-the-art RL-finetuned baselines on both general and domain-specific code editing. 
As shown in \cref{tab:main_result_editbench} and \cref{tab:main_result_kernelbench}, \sys outperforms the baselines across all tasks and models. 
On KernelBench, \sys outperforms the state-of-the-art Kevin-32B~\cite{baronio2025kevin} by 43.1\% in \emph{Correctness best@16} and achieves 1.1$\times$ in \emph{Speedup best@16} and \emph{avg@16}.
On EditBench, \sys with QwQ-32B matches the commercial gpt-o4-mini and significantly outperforms open-weight models (72B) by up to 6.9\% in \emph{Pass@1}.

\subsection{Ablations}
\label{subsec:eval_ablations}
We ablate each design component in \sys:
(1) the generative code representation learning, 
(2) the TTS that enforces semantics representation exploration before instruction-specific transformation, 
and (3) their combination.
\cref{tab:ablation_study} show the results. 

Specifically, we consider the following variants to isolate the contribution of each component: 
(1) \emph{Baseline}: a model finetuned only on instruction-following data combined with the TTS that directly applies instruction-specific transformation, without incorporating semantics-preserving transformation in either training or inference;
(2) \emph{+\sys TTS}: the baseline finetuned model combined with \sys TTS where explore semantics-preserving transformation is enforced;
(3) \emph{+\sys-trained}: a model trained with generative code representation learning and then finetuned, combined with the TTS used in the baseline,
(4) \emph{+Both}: the \sys-trained model combined with \sys TTS, representing the full \sys approach. 

Notably, \emph{\sys-trained} model uses \emph{strictly the same training budget} as the \emph{finetuned} baseline, e.g., the combined training iterations for both generative representation learning and finetuning equal to that of the baseline's finetuning iterations, to ensure a fair comparison.

\para{Generative code representation learning.}
Comparing \emph{Baseline} to \emph{+\sys-trained} in \cref{tab:ablation_study}, generative code representation learning alone provides consistent improvement across all metrics by 21\% in \emph{Speedup avg@16} and 1.05$\times$ in \emph{Correctness avg@16} on KernelBench and 3.5\% in \emph{Pass@1} and 13.32\% in \emph{Pass@16} on EditBench.

\para{\sys TTS.}
Comparing \emph{Baseline} to \emph{+\sys TTS} in EditBench, \sys TTS consistently improve \emph{Pass@1} and \emph{Pass@16} by 5.18\% and 10.45\% on EditBench, respectively.
In KernelBench, \sys TTS alone improves 10.5\% in \emph{Speedup best@16}.
While \emph{Correctness best@16} drops significantly, this can be attributed to the untrained model producing low-quality representations, which leads to incorrect exploration and thus, the wasted computational budget.

\begin{table*}[!t]    
    \centering
    \setlength{\tabcolsep}{5pt}
    \renewcommand{\arraystretch}{1}
    \small
    \caption{Comparing \sys to~\citet{baronio2025kevin} against unseen hardware devices, i.e., H200, for generalization by measuring performance on KernelBench~\citep{ouyang2025kernelbench}.}
    \label{tab:cross_device_generalization_kernelbench}
    \begin{tabular}{rllllllll}
    \toprule
    \multirow{2}{*}{} &\multicolumn{2}{c}{Correctness$\uparrow$} &\multicolumn{2}{c}{Speedup$\uparrow$} &\multicolumn{2}{c}{fast$_1\uparrow$}&\multicolumn{2}{c}{fast$_{1.5}\uparrow$}  \\
    \cmidrule{2-9}
     Model & best@16 & avg@16 & best@16 & avg@16 & best@16 & avg@16 & best@16 & avg@16\\
     \midrule
     Kevin-32B & 56 & 10.81  & 1.51 & 1.14 & 10 & 1.88 & 6 & 1.5 \\
     Finetuned QwQ-32B & 61$_{+5}$ & 12.63$_{+1.82}$ & \textbf{1.7}$_{+0.19}$ & 1.13$_{+0.01}$ & 15$_{+5}$ & 3.94$_{+2.06}$ & 8$_{+2}$ & 2.08$_{+0.58}$ \\
     \rowcolor{purple!8} \textbf{\sys (Ours)}  & \textbf{81}$_{+25}$ & \textbf{20.38}$_{+9.57}$ &  1.62$_{+0.11}$ &\textbf{1.25}$_{+0.11}$ & \textbf{17}$_{+7}$ & \textbf{4.13}$_{+2.25}$ & \textbf{8}$_{+2}$ & \textbf{2.25}$_{+0.75}$ \\  
    \bottomrule
    \end{tabular}
\end{table*}

\begin{table}[!t]
    \centering
    \small
    \setlength{\tabcolsep}{8pt}
    \renewcommand{\arraystretch}{1}
    \small
    \caption{Comparing \sys-trained model to state-of-the-art models against semantics-preserving code transformations.}
    \label{tab:robustness_editbench}
    \begin{tabular}{r|ll}
        \toprule
        Model &  Pass@1 $\uparrow$ & Consistency $\uparrow$ \\
        \midrule
        QwQ-32B                        &     47.22 & 76.85\\
        Finetuned QwQ-32B             &     50.93$_{+3.71}$ & 75.00$_{-1.85}$\\
        GPT o4-mini                    &     53.70$_{+6.48}$  & 83.33$_{+6.48}$\\
        \rowcolor{purple!8} \textbf{\sys (Ours)}  & \textbf{53.70$_{+6.48}$} & \textbf{88.89$_{+12.04}$} \\
        \bottomrule
    \end{tabular}
\end{table}

\para{Full \sys.}
When turning on both training and inference in \sys, it outperforms \emph{Baseline} by 2$\times$ and 1.2$\times$ in \emph{Correctness avg@16} and \emph{Speedup best@16} on KernelBench, and 7\% and 20\% in \emph{Pass@1} and \emph{Pass@16} on EditBench, respectively.
This demonstrates that the training and inference components are designed to work together effectively.

\subsection{Robustness and Generalization}
\label{subsec:eval_robustness_generalization}

Real-world code editing must operate reliably across diverse code styles, formats, and syntactic structures.
Models that overfit to surface-level syntax will fail when encountering code that deviates from their training distribution.
This section evaluates whether \sys's explicit training on semantics-preserving transformations improves the model's robustness and generalization to syntactic changes.
For example, the \sys's transformations could potentially canonicalize the syntactically perturbed code during inference, or help the model reason about the inherent equivalence of the perturbed code via training.
To this end, we measure robustness and generalization across two dimensions: (1) robustness to semantics-preserving code perturbations, and (2) generalization to unseen hardware.

\para{Robustness.}
We compare \sys to the state-of-the-art baseline against semantics-preserving code transformation.
Following ReCode~\cite{recode_wang2022}, we adopt three main types of perturbations: NLAugmenter~\cite{dhole2021nlaugmenter} for docstring, NatGen~\cite{chakraborty2022natgen} for code syntax, and code format.
However, KernelBench input mainly consists of PyTorch API calls, leaving limited room for these perturbations.
Therefore, we evaluate robustness only on EditBench.

As shown in \cref{tab:robustness_editbench}, we report (1) \textit{Pass@1} after the testing samples are perturbed, and (2) \textit{Consistency}, which measures the proportion of samples have maintained their results after the perturbation, i.e., the pass/fail outcome of each test cases is preserved.
\sys remains robust and less susceptible than \emph{Baseline}, achieving 6.7\% better consistency and remains the highest \textit{Pass@1} after perturbations.

\para{Cross-device generalization.} 
Low-level CUDA optimizations (e.g., shared memory tiling, warp-level primitives) are highly sensitive to hardware architecture.
We investigate \sys's generalizability to unseen devices when performing low-level domain-specific CUDA optimization.
To evaluate whether the trained models overfit to the training hardware, we evaluate on unseen H200 devices without any additional training.
\cref{tab:cross_device_generalization_kernelbench} demonstrates that \sys generalizes effectively to unseen devices, outperforming \emph{Baseline} (Finetuned QwQ-32B) by 10.6\% in \emph{Speedup avg@16} while maintaining 61.4\% higher in \emph{Correctness avg@16}.

\subsection{Test Time Scaling}
\label{subsec:eval_test_time_scaling}

\begin{figure}[!t]
    \centering
    \includegraphics[width=0.95\linewidth]{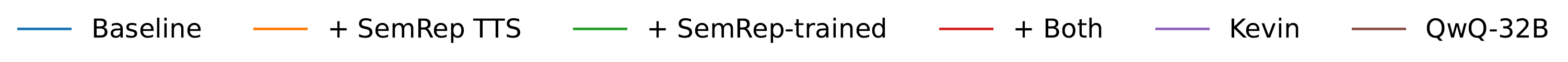}
    \includegraphics[width=0.48\linewidth]{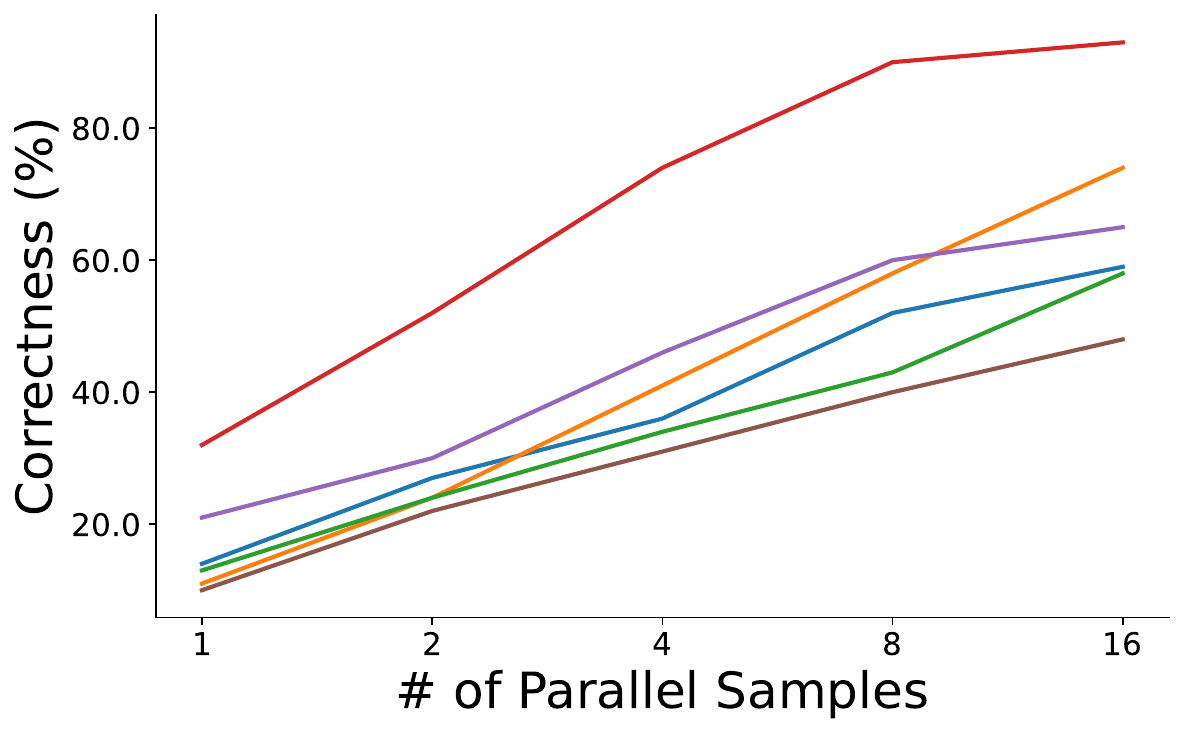}
    \includegraphics[width=0.48\linewidth]{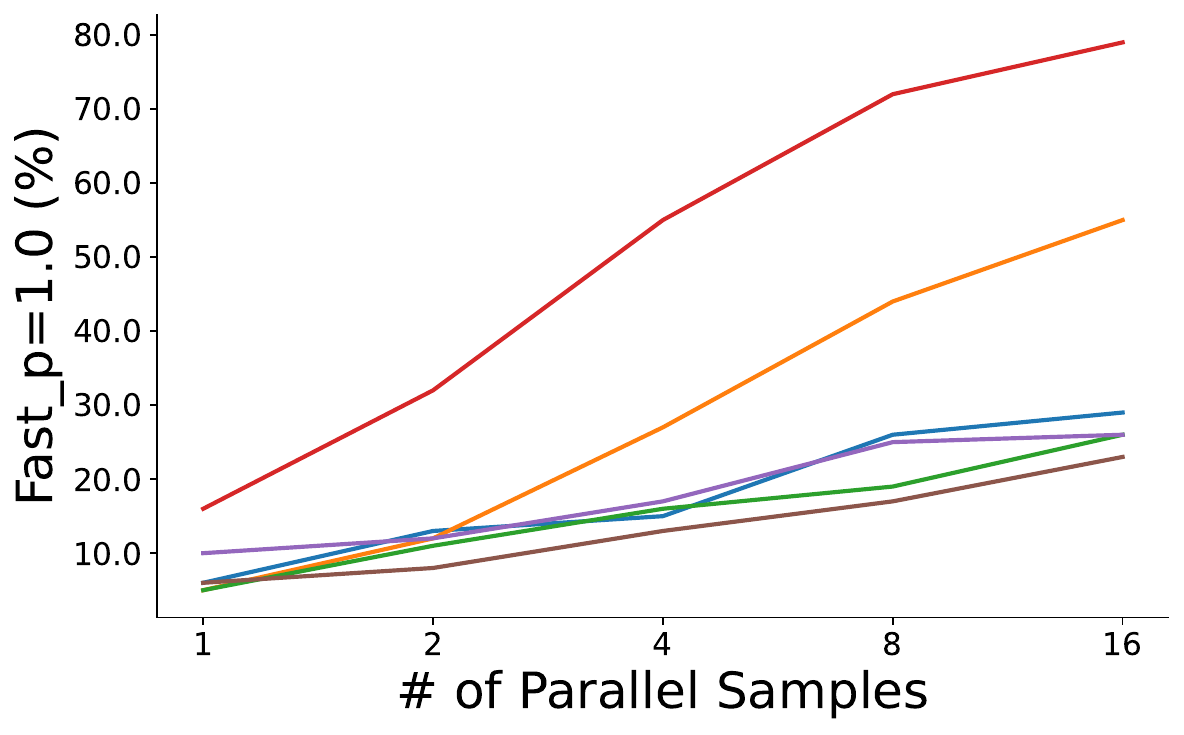}
    \caption{Scaling the number of parallel samples on KernelBench.}
    \label{fig:scaling_parallel_samples}
\end{figure}

As \sys produces explicit code representations as semantics-preserving programs during inference, it naturally supports test-time scaling, where additional compute is used to balance exploration (exploring various equivalent code implementations) and exploitation (sticking to narrow but effective transformations).
For example, AlphaEvolve~\cite{novikov2025alphaevolve} scales search by exploring semantically equivalent implementations of the same algorithm, while prior program repair works like REx~\cite{tang2024code} also formulate the iterative program refinement as explore–exploit trade-offs.
By exposing semantic structure rather than internalizing it in weights, \sys's output serves as meaningful stepping stones for later search.

\begin{figure*}[!t]
    \centering
    \subfloat[Direct transformation.]{
        \includegraphics[width=0.45\linewidth]{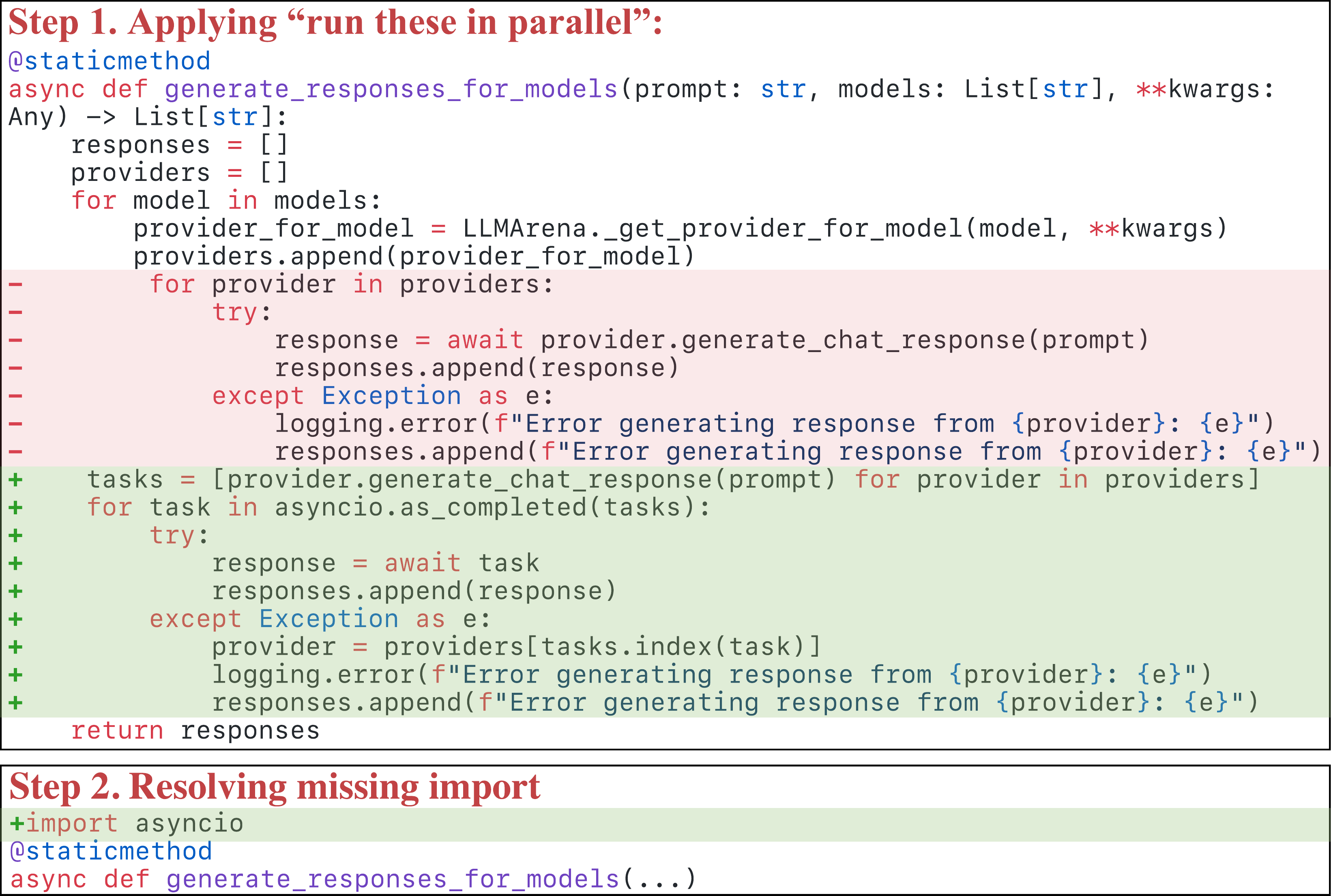}
    \label{fig:editbench-case-study-direct}
    }    
    \subfloat[\sys.]{
    \includegraphics[width=0.47\linewidth]{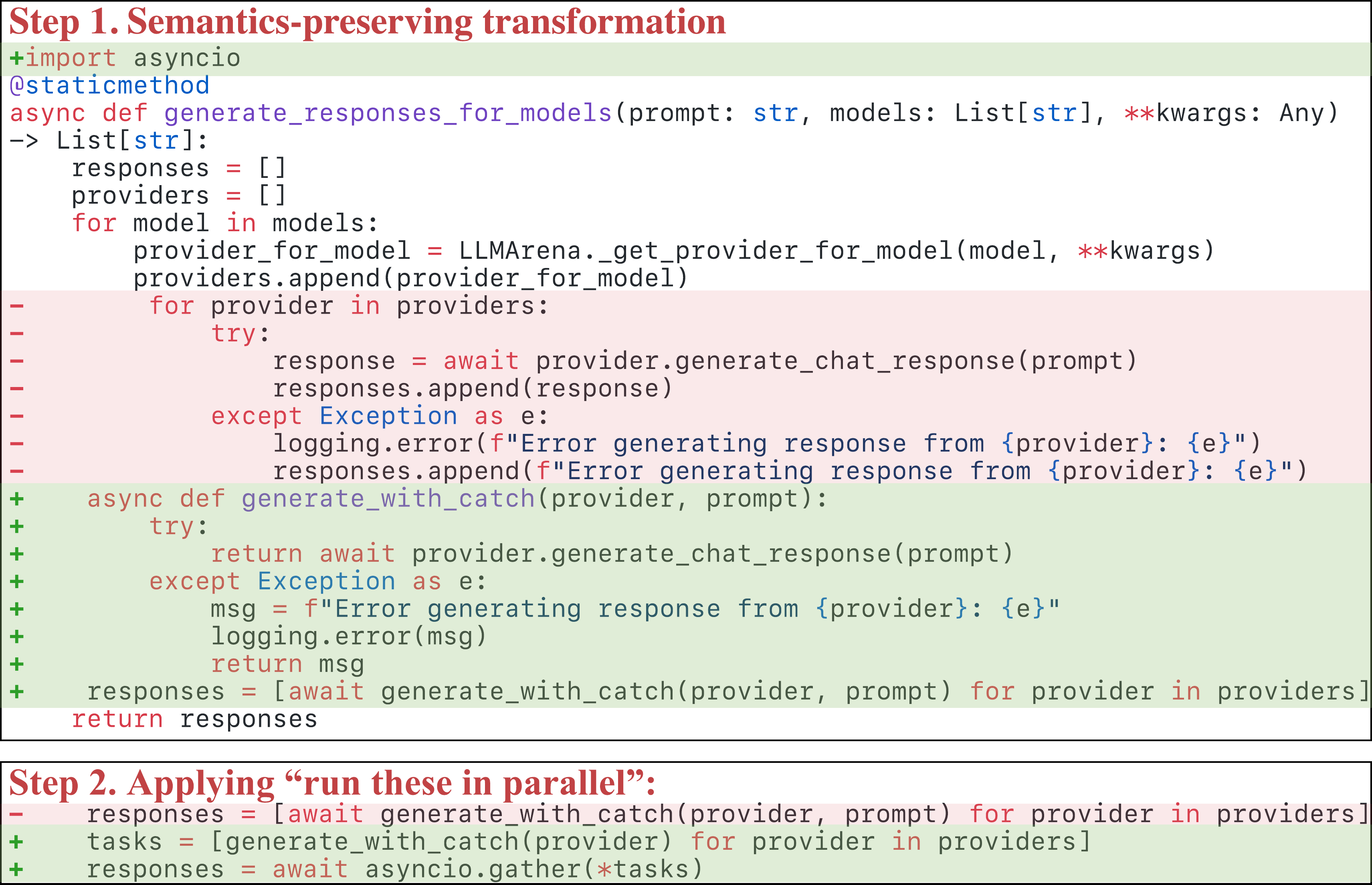}
    \label{fig:editbench-case-study-semrep}
    }
    \caption{An example in EditBench showing how (a) direct transformation and (b) \sys generate with ``run these in parallel''.}
    \label{fig:editbench-case-study}
\end{figure*}

To evaluate the amenability of \sys to test-time scaling, we study how \textit{best@k} performance scales when increasing the number of parallel trajectories $k$, while fixing the number of refinement turns.
We compare against the base QwQ-32B, the state-of-the-art Kevin-32B~\cite{baronio2025kevin}, and all variants from \cref{subsec:eval_ablations}. 

As shown in \cref{fig:scaling_parallel_samples}, \sys grows faster than all the baselines and variants when the number of parallel $k$ increases.
At $k=16$, \sys achieves up to 3$\times$ gains over \emph{Baseline} (Finetuned QwQ-32B) in \emph{fast$_{1.0}$} \emph{best@16}, while maintaining 4$\times$ higher in \emph{Correctness best@16}.
\subsection{Case Study on EditBench}
\label{subsec:case_study}
As shown in \cref{fig:editbench-case-study}, we examine how direct transformation and \sys handle a real-world code editing task from EditBench that requires implementing parallelism. 
Full trajectories are presented in Appendix~\ref{app:editbench_case_study}.
The original code (see \cref{lst:editbench_input} for the complete version) takes a prompt and a list of LLM models, calls each model's API sequentially via \texttt{generate\_chat\_response}, and returns a list of outputs in the same order as the input models.
Each call must complete before the next begins, so the total execution time is the sum of per-model response times.

\para{Direct transformation.}
Direct transformation (\cref{fig:editbench-case-study-direct}) conflates rewriting and optimization at each step, which makes it harder to get both right simultaneously.
Therefore, in \emph{Step 1}, it attempts to introduce parallelism by converting the sequential loop into concurrent API calls using \texttt{asyncio.as\_completed}, an async API that returns results as soon as each call finishes.
However, it omits the required import for \texttt{asyncio}.
Upon the external execution feedback, \emph{Step 2} attempts to fix this compilation error rather than making any meaningful progress.
While \emph{Step 2} fixes the issue, it breaks the semantics of output ordering even though the code runs faster.
The original code appends outputs in the order they are received as input, whereas \texttt{asyncio.as\_completed} returns results in completion order rather than the same order as the input models.
As a result, the $i$-th output list element no longer reliably matches the $i$-th model in the input list.

\para{\sys.} 
\sys (\cref{fig:editbench-case-study-semrep}) decouples the semantic-preserving refactoring required for the optimization from the optimization itself by separating them into two distinct steps.
In \emph{Step 1}, \sys is prompted to produce a semantic equivalent but not necessarily optimized code.
As a result, \sys shuffles code around and extracts a helper function \texttt{generate\_with\_catch} to collect the API responses.
Note that this transformation does not optimize the performance but abstracts the per-call logic into a self-contained module, making the parallelism opportunity more explicit in structure.
In \emph{Step 2}, \sys is prompted to follow the user's instructions to optimize the refactored code from \emph{Step 1}.
With the helper function in place, \sys can focus on implementing parallel execution of the API calls.
Specifically, \sys uses the async API \texttt{asyncio.gather}, which concurrently runs the helper function for all models and returns the responses in the same order as the input models.
By separating the required semantics-preserving refactoring (\emph{Step 1}) from the instruction following (\emph{Step 2}), \sys avoids the correctness pitfalls that potentially arise when both goals compete within a single generation.

\subsection{Integrating \sys with Evolutionary Agent}
\label{subsec:eval_evolve_agent}

As \sys is explicitly trained on semantics-equivalent transformations, it has the potential to produce diverse intermediate programs not yet optimized, but potentially include rewrites helpful to enable future optimizations.
This behavior aligns well with evolutionary search, which also benefits from exploring non-greedy intermediate states.
Therefore, we embed \sys into the evolutionary agentic framework OpenEvolve~\cite{openevolve} for code optimizations.
All compared models, i.e., DeepSeek-V3-Chat and DeepSeek-V3-Reasoner, are run within OpenEvolve under the same agent configurations.
See Appendix~\ref{app:openevolve} for implementation details.

\para{Minimum spanning tree (MST).}
We adopt the minimum spanning tree task from AlgoTune~\cite{press2025algotune} for the case study.
This problem seeks to find a subset of edges that connects all vertices in a weighted, undirected graph with the lowest possible total edge weight and no cycles.

\begin{figure}[!t]
    \centering
    \includegraphics[width=\linewidth]{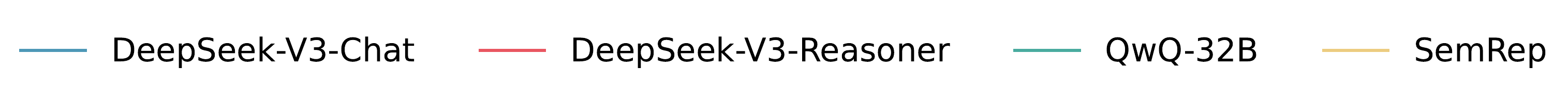}
    \subfloat[Minimum Spanning Tree]{
        \includegraphics[width=0.46\linewidth]{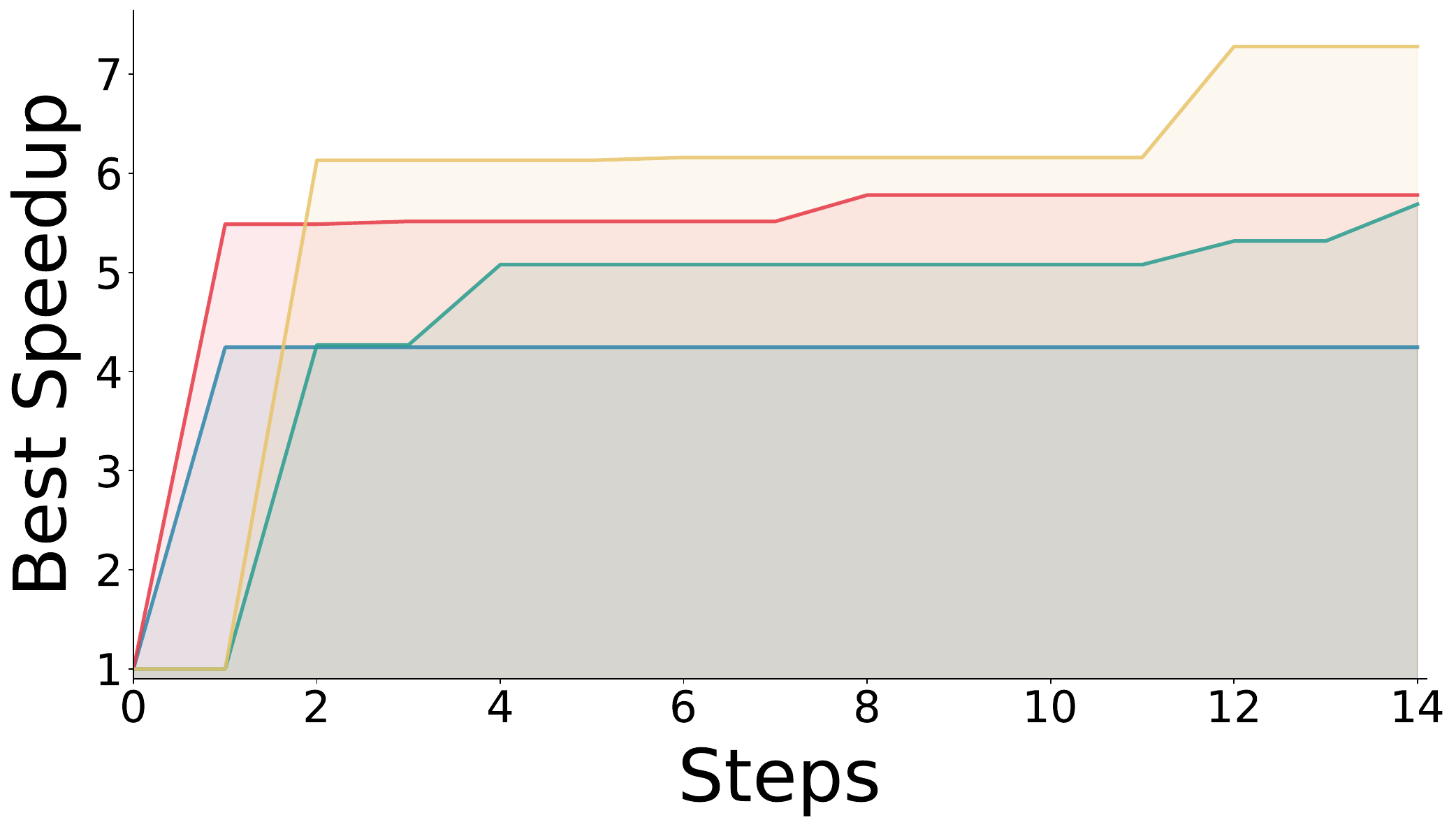}
    \label{fig:openevolve_mst_results}
    }
    \subfloat[Circle Packing]{
        \includegraphics[width=0.46\linewidth]{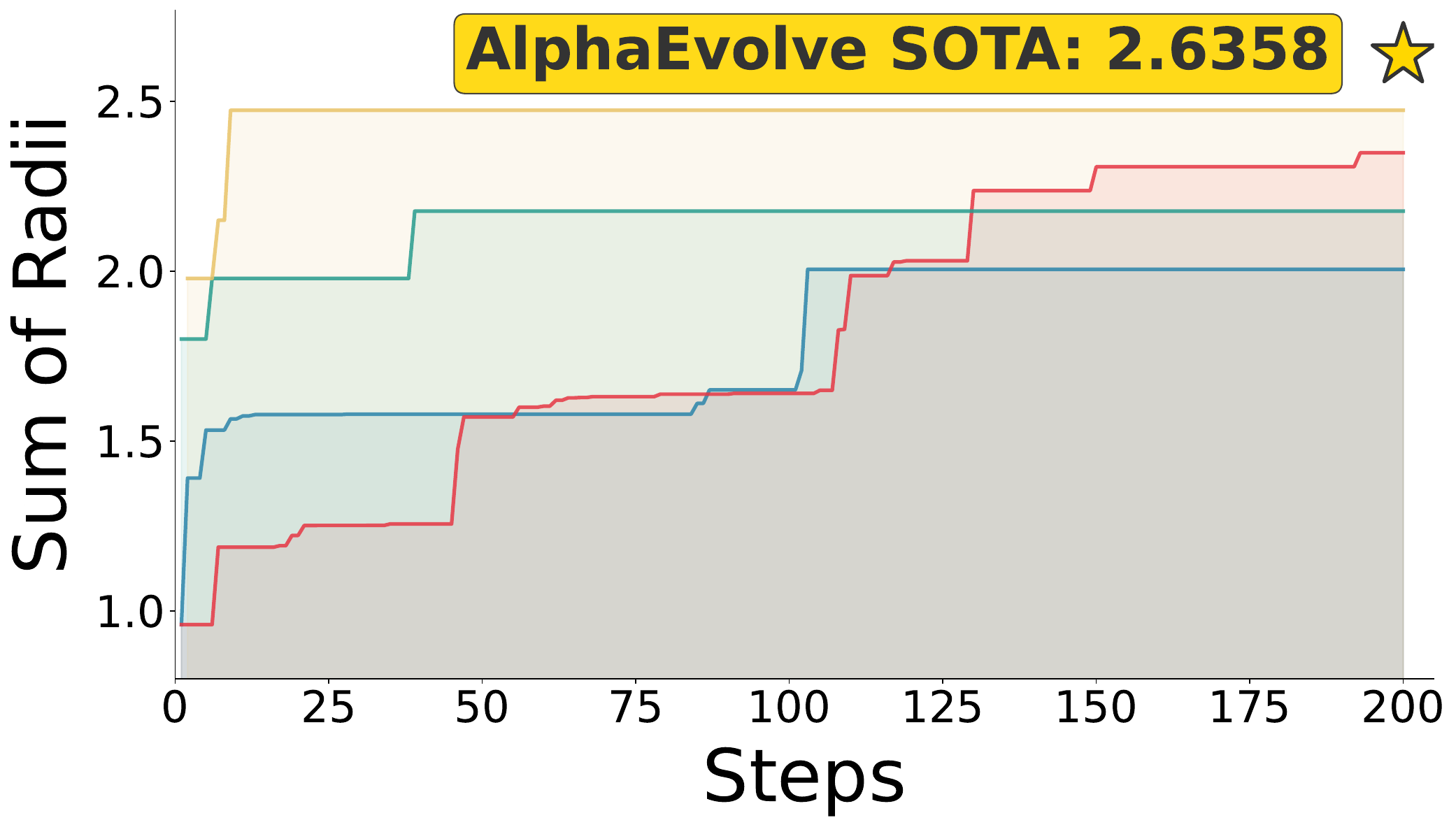}
    \label{fig:openevolve_circle_packing_results}
    }
    \caption{Best results over optimization steps with OpenEvolve for the minimum spanning tree and circle packing problem.}
\end{figure}

As shown in \cref{fig:openevolve_mst_results}, \sys achieves a speedup of 7.28x, outperforming both 685B DeepSeek-V3-Chat and DeepSeek-V3-Reasoner by 25.95\% at step 14.
Starting from the initial NetworkX~\cite{hagberg2007exploring} implementation (\cref{lst:mst_checkpoint_0}), besides replacing NetworkX with a custom Kruskal's algorithm, \sys achieves a significant 6.13$\times$ speedup through implementing the Union-Find algorithm in an iterative way, which all the other models fail to identify (see \cref{lst:mst_semrep} and \ref{lst:mst_others}).
Ultimately, \sys achieves the final 7.28$\times$ speedup by applying micro-optimizations, including \texttt{operator.itemgetter} for sorting and precomputation to improve cache locality (\cref{lst:mst_checkpoint_12}).
See Appendix~\ref{app:openevolve_mst} for detailed analysis.
\begin{lstlisting}[language=Python, basicstyle=\tiny, label=lst:mst_semrep, caption=\sys implements the Union-Find algorithm with efficient while loop.,]
def find(u: int) -> int:
    while parent[u] != u: 
        # Path compression
        parent[u] = parent[parent[u]] 
        u = parent[u]
    return u
\end{lstlisting} 
\begin{lstlisting}[language=Python, basicstyle=\tiny, label=lst:mst_others, caption=All the other models implement Union-Find with recursion.,]
def _find(self, parent: List[int], x: int) -> int:
    """Find with path compression."""
    if parent[x] != x:
        parent[x] = self._find(parent, parent[x])
    return parent[x]
\end{lstlisting}

\para{Circle packing.}
We adopt the circle packing problem studied in AlphaEvolve~\cite{novikov2025alphaevolve}.
The circle packing problem is a constrained optimization task that seeks to arrange $n=26$ non-overlapping circles within a unit square to maximize the sum of their radii.

As shown in \cref{fig:openevolve_circle_packing_results}, integrating \sys into OpenEvolve achieves the sum radii of 2.4737 (93.88\% of the state-of-the-art AlphaEvolve results). 
Notably, AlphaEvolve leverages Gemini models over many thousands of iterations, while our \sys achieves competitive performance under substantially more constrained resources.
Moreover, it still significantly outperforms OpenEvolve paired with 685B DeepSeek-V3-Chat and Reasoner models~\cite{deepseekai2024deepseekv3technicalreport}.
Through evolutionary search, \sys discovers an effective grid-based pattern in the early stage, and then switches to an alternative ring-based approach that temporarily reduces the performance. 
Although this semantics-preserving change does not yield immediate improvement and even harms the performance, the algorithmic shift opens up new opportunities for further optimization.
Ultimately, at step 93, \sys achieves the improvement by starting from a ring-based initialization that provides a favorable prior, and then applies the simulated annealing to relocate components for better performance.
See Appendix~\ref{app:openevolve_circle_packing} for details.

\section{Discussion and Limitations}
\label{sec:discussion}
\para{Test availability.}
\sys's definition of equivalence is defined over a fixed set of inputs.
This is subject to the availability of tests for computing verifiable rewards during representation learning, and lacks a soundness guarantee.
If these tests have low coverage, i.e., the inputs do not exercise most of the code, the \sys-trained model may generate biased code that is only semantically equivalent on a small part of the code logic.
Exploring stronger equivalence oracles, such as symbolic execution, would be an interesting direction for future work.

\para{Reward hacking.}
During the generative code representation learning, we reject exact duplicates to prevent reward hacking, where a model could trivially return the input code unchanged to maximize the reward.
However, minor surface-level refactorings, e.g., adding comments, renaming variables, are deliberately allowed.
These lightweight transformations may help the model understand code behavior, building a richer representation that can, in turn, inspire more high-quality transformations at inference, such as performance optimization and functional adaptation to new requirements.
Determining which semantic-preserving transformations are most beneficial for model training remains an open question.
It is also challenging to directly measure how the explicit code representation learning encourages more exploration and leads to more creative transformations, without resorting to the outcomes from the downstream editing tasks.
We are working on including additional categories of useful semantics-preserving transformations beyond those presented in this paper.

\para{Similarity to curriculum learning.}
Our scheduling of the reward for representation learning and finetuning can also be made continuous as a curriculum, i.e., start from a high reward on generating semantically equivalent code, and gradually decrease it to prioritize instruction-specific transformation.
The key advantage of \sys's strict two-stage training is that its mid-training is a one-time effort that facilitates many downstream editing tasks, rather than being specialized in a single task-specific setting, which can suffer from limited generalization.

\para{Evolutionary agentic framework.}
While \sys utilizes an iterative inference paradigm, the current implementation is largely simplified due to computational constraints.
This setup may not fully unleash the potential of the \sys's trained model for challenging coding tasks that require many steps to solve.
While our case studies on integrating \sys with evolutionary coding agents demonstrate promising results, we leave a systematic study and broader application to real-world projects for future work.
\section{Related Work}
\label{sec:related_work}
\para{Code editing and transformation.}
Code editing has emerged as a critical application of LLMs in software engineering, with applications spanning code maintenance~\cite{chi2025editbenchevaluatingllmabilities, jimenez2023swebench, wei2025swe}, performance optimization~\cite{pie_iclr_2024_spotlight, huang2024effi, peng2025perfcodegen, garg2022deepdev, press2025algotune, shetty2025gso}, bug fixing~\cite{xia2024automated, jin_inferfix_2023}, code migrations~\cite{liu_migrationbench_2025, zhang_scalable_2024}.
Most existing approaches either adopt finetuning to learn a direct mapping between code pairs or leverage iterative refinement with execution or self-generated feedback~\cite{huang2024effi, peng2025perfcodegen, huang2023agentcoder, xia2024automated, chen2023teaching, dong2024self, madaan2024self, zelikman2023self, liu2024learning}.
While these methods have shown effectiveness, they treat the semantic understanding as an implicit byproduct of the editing task rather than an explicit learning objective.
In contrast, \sys complements these approaches by treating semantics-preserving transformation as an explicit foundational capability, explicitly grounding transformation reasoning in verifiable representations that benefit both training and inference.

\para{Code semantics learning.}
Previous works have attempted to capture code semantics through execution-aware learning~\cite{pei2020trex, ni_next_2024, liu_code_2023, ding_semcoder_2024} or structural constraints like program dependence graphs~\cite{pei2023exploiting, guo2020graphcodebert}.
However, these methods are limited to specific program representations with nontrivial construction costs and limited generality.
\sys formulates semantic understanding as a generative task where the model explicitly produces semantically equivalent code as a verifiable intermediate representation, making the semantic understanding amenable to reinforcement learning.

\para{Reinforcement pre-training and evolutionary search.}
\sys shares a similar spirit to reinforcement pre-training~\cite{hatamizadeh_rlp_2025, dong_reinforcement_2025, copet2025cwm}, where the exploration is encouraged and introduced during pre-training before task-specific training to improve generalization.
\sys also relates to evolutionary agentic frameworks~\cite{openevolve,novikov2025alphaevolve,tang_code_2024}, where inference is designed to balance exploration and exploitation~\cite{tang_code_2024}.
\sys extends the idea by explicitly disentangling the representation learning from instruction-specific transformation, enabling LLMs to leverage improved semantic understanding for improved performance and generalization (\cref{subsec:eval_robustness_generalization}).
\section{Conclusion}
\label{sec:conclusion}
We introduce \sys, a code transformation framework that decouples semantic understanding from code editing.
Our key approach is to employ \emph{generative code representation learning} to enable models to reason about program behavior as explicit representations.
\sys enables smaller models to match or exceed larger models, achieving significant gains across both general and domain-specific code editing tasks, while exhibiting enhanced robustness to semantics-preserving transformations and improved generalization to unseen hardware.


\bibliography{references}
\bibliographystyle{icml2026}

\newpage
\appendix
\onecolumn

\appendix
\section{Experiment Details}
\subsection{Prompt Details}
\begin{PromptBox}{Prompt for semantic-preserving transformation on KernelBench.}
\label{prompt:semantic-transformation}
You are given the following architecture:\\
\texttt{\{ref\_arch\_src\}}\\
\\
Perform a semantic-preserving transformation on the given architecture. This means you should rewrite the code in a way that:\\
- Retains the exact behavior and output of the original code\\
- Changes the style, structure, algorithm, or abstraction level\\
- Improves the code structure without changing its semantics\\
- You are free to be creative with the transformation while ensuring semantic equivalence\\
\\
Use \texttt{torch.utils.cpp\_extension.load\_inline} and name your optimized output architecture \texttt{ModelNew}. Output the new code in code blocks. Please generate real code, NOT pseudocode, and make sure the code compiles and is fully functional. After your answer, summarize your changes in a few sentences.
\end{PromptBox}

\begin{PromptBox}{Prompt for instruction-specific transformation on KernelBench.}
\label{prompt:cuda-optimization}
You are given the following architecture:\\
\texttt{\{ref\_arch\_src\}}\\

Optimize the architecture named \texttt{Model} with custom CUDA operators! Use techniques like:\\
- Shared memory for tile-based computation\\
- Coalesced memory access\\
- Warp-level primitives\\
- Tensor cores if applicable\\
\\
Use \texttt{torch.utils.cpp\_extension.load\_inline} and name your optimized output architecture \texttt{ModelNew}. Output the new code in code blocks. Please generate real code, NOT pseudocode, and make sure the code compiles and is fully functional. After your answer, summarize your changes in a few sentences.
\end{PromptBox}

\begin{PromptBox}{Prompt for semantic-preserving transformation on EditBench.}
You are given the following code to modify:
\\
The Original code (to be modified):\\
\\
\texttt{\{lang\}}\\
\texttt{\{original\_code\}}\\
\\
Perform a semantic-preserving transformation on the code that: \\
- Maintains the exact behavior and output for the rest of the code \\
- Uses semantic-preserving transformations where possible (e.g., restructuring, renaming variables, using equivalent operations) \\
- Preserves all side effects, error behaviors, and corner-case handling \\
\\
Please output the entire modified code file in a code block beginning with ```\texttt{\{lang\}}'''. After your answer, summarize your changes in a few sentences.
\end{PromptBox}

\begin{PromptBox}{Prompt for instruction-specific transformation on EditBench.}
\label{prompt:code-modification}
Generate a new implementation of the following code based on the user instruction:\\
\\
The Original code (to be modified):\\
\\
\texttt{\{lang\}}\\
\texttt{\{original\_code\}}\\
\\
The user instruction is:\\
\texttt{\{instruction\}}\\
\\
And they highlighted this section to be changed:\\
\texttt{\{lang\}}\\
\texttt{\{highlighted\_code\}}\\

Please only change the highlighted section and leave the rest of the code unchanged.\\
Please output the entire code file.\\
Respond only in a code block beginning with \texttt{\{lang\}}.
\end{PromptBox}
\subsection{Dataset and Hyperparameters}
\label{app:dataset_hyperparameters}
\paragraph{Dataset.}
For EditBench, we train on 82 randomly selected CodeContests samples~\cite{codecontest,pie_iclr_2024_spotlight} and the EDIT-Bench-complete dataset (excluding the core test set) and evaluate on the core set.
Since EditBench does not distinguish between tests for preserved functionality and intended new functionality, we manually run the provided tests on the input code. 
Tests that the original code can pass are considered as semantic preservation verification ($\mathcal{U}_{sem}$), while tests that the original code fails to pass (but the expected edited code should pass) are treated as instruction adherence verification ($\mathcal{U}_{edit}$).

For KernelBench, following Kevin~\citep{baronio2025kevin}, we randomly select 180 tasks from the benchmark for training and use the remaining 100 for testing.
During evaluation, we also identified and fixed a bug in the KernelBench speedup calculation.
Specifically, the original implementation did not enforce consistent sorting by \textit{problem\_id} before performing pair-wise speedup comparisons.
Thus, when evaluation results were incomplete, this inconsistency led to mismatched pairs and statistically incorrect speedup ratios.

\paragraph{Hyperparameters.}
We set the maximum training sequence length to 16,384 tokens to accommodate long reasoning trajectories.
The learning rate is set to 2e-6, and the batch size is 8. 
We set $\alpha_1=1.3, \beta_1=0.5$ for generative code representation learning, and $\alpha_2=1, \beta_2=0.3, \gamma=0.1$ for instruction-specific transformation learning.
During inference, we set the maximum sequence length to 16,384 tokens, with temperature 0.7 and top\_p=0.9 to balance exploration and exploitation.
We use $\omega_1=1, \omega_2=0$ for semantics-preserving transformation, and $\omega_1=0.3, \omega_2=0.7$ for instruction-following transformation.

\subsection{Budget}
\label{app:budget}
In order to achieve a fair comparison, we make sure the \sys-trained model uses strictly the same training budget compared to the extensively finetuned baselines.
We consider the total training budget $\mathcal{B}$ as the total number of training steps used by \emph{Baseline}.
For \sys, we take $\mathcal{B}/2$ steps for representation learning, and the other $\mathcal{B}/2$ steps for instruction-specific learning.

\subsection{Things We Tried that Did not Work}
We experimented with several alternative ways to learn code representations, but found that they underperformed.
\paragraph{Separate samples for distinct learning phases.}
Generative code representation learning and instruction-specific transformation learning should be conducted on different data samples to prevent conflicting learning signals.
Mixing or overlapping the same instances across both phases led to gradient conflicts and degraded performance. 
\paragraph{Stable learning objectives.}
The learning objective (generative code representation vs.\ instruction-specific transformation reward) should not be switched too rapidly during training.
Frequently alternating objectives caused the evolving rewards problem, where the stale training experiences become misleading, resulting in degraded performance.

\section{OpenEvolve Case Study}
\label{app:openevolve}
This section provides implementation details for integrating our \sys-trained QwQ-32B into OpenEvolve~\cite{openevolve}. 
We demonstrate the integration using two case studies: the minimum spanning tree task from AlgoTune~\cite{press2025algotune} and the circle packing problem.

\subsection{AlgoTune}
\label{app:openevolve_mst}
\paragraph{Hyperparameters.}
The evolutionary search was configured with a population size of 1000 and an archive size of 100, distributed across 4 islands to promote diversity. To balance search dynamics, we set the elite selection ratio to $0.1$, exploration ratio to $0.3$, and exploitation ratio to $0.6$. For island communication, we employed a migration rate of $0.1$. During prompt construction, we included the top 3 programs and 2 diverse programs. The evolution was run for a maximum of 16 iterations. The LLM was configured with a temperature of $0.7$, top-$p$ of $0.9$, and a maximum generation length of 16,384. The \texttt{random\_seed} was 42, and \texttt{diff\_based\_evolution} was set to false.

\paragraph{Evaluator modifications.}
We refined the evaluation metric to better align with the goal of performance optimization. The original AlgoTune evaluator calculated a composite score as follows:
\begin{equation}
    \text{score} = 0.7 \cdot \text{Correctness} + 0.2 \cdot \text{Performance} + 0.1 \cdot \text{Reliability}
\end{equation}
In this formula, \textit{Reliability} measures the ratio of successful executions to total trials. \textit{Correctness} represents the proportion of valid solutions among successful runs, and \textit{Performance} is calculated as $1/(1+t)$, where $t$ denotes the execution time in milliseconds. We found that the absolute runtime dependency in the performance term introduced variance due to baseline fluctuations, making it an unstable proxy. Consequently, we modified the evaluator (e.g., in \texttt{minimum\_spanning\_tree/evaluator.py}) to define the fitness score directly as the \textit{speedup} ratio relative to the baseline implementation for correct solutions, while assigning a score of zero if incorrect. This ensures that the evolutionary process directly optimizes for relative efficiency gains unaffected by environmental noise.
All compared methods were re-evaluated under this modified evaluator.

\paragraph{Trajectory analysis.}
Starting from the initial NetworkX implementation, \sys evolves to achieve a 7.28$\times$ speedup, outperforming both QwQ-32B and DeepSeek-V3-Reasoner.
We track the evolution trajectories of \sys in detail, examining important breakthroughs at each stage of its evolution:
\begin{itemize}
    \item \textbf{Initial implementation (\cref{lst:mst_checkpoint_0}).}
    The initial implementation uses the NetworkX library~\cite{hagberg2007exploring} for graph construction and MST computation.
    The implementation suffers from significant library overhead, complex graph data structures, and multiple data format conversions.

    \item \textbf{Custom algorithm implementation (\cref{lst:mst_checkpoint_2}).}
    \sys replaces the NetworkX-based implementation with a custom MST implementation.
    The key changes that lead to the performance breakthrough include: (1) elimination of NetworkX dependency and library overhead, (2) direct implementation of Kruskal's algorithm with greedy edge selection, (3) an iterative Union-Find data structure, and (4) early exit optimization once the MST is complete.
    This implementation represents a 6.13$\times$ speedup over the baseline, demonstrating \sys's ability to recognize performance bottlenecks and discover efficient algorithms.

    \item \textbf{Stabilization (\cref{lst:mst_checkpoint_4}).}
    \sys maintains the optimized algorithm while improving code quality with type hints and better documentation.
    Performance remains stable at around 6$\times$ speedup.

    \item \textbf{Final micro-optimizations (\cref{lst:mst_checkpoint_12}).}
    \sys applies micro-optimizations that improve cache locality and reduce function call overhead, achieving the final 7.28$\times$ speedup.
    The key optimizations include: (1) using \texttt{operator.itemgetter} instead of lambda functions for sorting, (2) precomputing the target value (\texttt{num\_nodes - 1}) to avoid repeated calculations in the tight loop, and (3) consistent use of C-implemented operator functions for all sorting operations.
    These micro-optimizations provide an additional 21\% improvement.
\end{itemize}

\begin{lstlisting}[language=Python, label={lst:mst_checkpoint_0}, caption=MST: initial implementation using NetworkX., firstline=44, lastline=104]
import logging
import random
import networkx as nx
import numpy as np
from typing import Any, Dict, List, Optional

class MinimumSpanningTree:
    """
    Initial implementation of minimum_spanning_tree task.
    This will be evolved by OpenEvolve to improve performance and correctness.
    """
    
    def __init__(self):
        """Initialize the MinimumSpanningTree."""
        pass
    
    def solve(self, problem):
        """
        Solve the minimum_spanning_tree problem.
        
        Args:
            problem: Dictionary containing problem data specific to minimum_spanning_tree
                   
        Returns:
            The solution in the format expected by the task
        """
        try:
            """
            Construct the graph in networkx, compute MST using minimum_spanning_edges,
            and return the MST edges sorted by (u, v).

            :param problem: dict with 'num_nodes', 'edges'
            :return: dict with 'mst_edges'
            """
            G = nx.Graph()
            num_nodes = problem["num_nodes"]
            edges = problem["edges"]

            G.add_nodes_from(range(num_nodes))
            for u, v, w in edges:
                G.add_edge(u, v, weight=w)

            # networkx returns an iterator of MST edges
            mst_edges_data = list(nx.minimum_spanning_edges(G, data=True))

            # Convert to [u, v, weight]
            # each edge is (u, v, {'weight': w})
            mst_edges = []
            for u, v, data in mst_edges_data:
                # ensure u < v for sorting consistency
                if u > v:
                    u, v = v, u
                mst_edges.append([u, v, data["weight"]])

            # Sort by (u, v)
            mst_edges.sort(key=lambda x: (x[0], x[1]))
            return {"mst_edges": mst_edges}
            
        except Exception as e:
            logging.error(f"Error in solve method: {e}")
            raise e
\end{lstlisting}
\begin{lstlisting}[language=Python, label={lst:mst_checkpoint_2}, caption=MST: custom Kruskal's algorithm with Union-Find., firstline=44, lastline=117]
import logging
from typing import Any, Dict, List, Optional

class MinimumSpanningTree:
    """
    Implementation of minimum_spanning_tree task using Kruskal's algorithm for improved performance.
    """
    
    def __init__(self):
        """Initialize the MinimumSpanningTree."""
        pass
    
    def solve(self, problem: Dict[str, Any]) -> Dict[str, List[List[float]]]:
        """
        Solve the minimum_spanning_tree problem using Kruskal's algorithm.
        
        Args:
            problem: Dictionary containing problem data specific to minimum_spanning_tree
                   
        Returns:
            The solution in the format expected by the task
        """
        try:
            num_nodes = problem["num_nodes"]
            edges = problem["edges"]

            # Sort edges by weight to facilitate Kruskal's algorithm
            edges_sorted = sorted(edges, key=lambda x: x[2])

            parent = list(range(num_nodes))
            size = [1] * num_nodes  # Using size for union by larger size

            def find(u: int) -> int:
                while parent[u] != u:
                    parent[u] = parent[parent[u]]  # Path compression
                    u = parent[u]
                return u
            
            def union(u: int, v: int) -> bool:
                u_root = find(u)
                v_root = find(v)
                if u_root == v_root:
                    return False  # Already connected
                # Union by size: attach smaller to larger
                if size[u_root] < size[v_root]:
                    parent[u_root] = v_root
                    size[v_root] += size[u_root]
                else:
                    parent[v_root] = u_root
                    size[u_root] += size[v_root]
                return True

            mst_edges = []
            edge_count = 0

            for edge in edges_sorted:
                u, v, w = edge
                if find(u) != find(v):
                    union(u, v)
                    # Ensure consistent ordering for output sorting
                    if u > v:
                        u, v = v, u
                    mst_edges.append([u, v, w])
                    edge_count += 1
                    if edge_count == num_nodes - 1:
                        break  # Early exit once MST is complete

            # Final sort by (u, v) for consistent output
            mst_edges.sort(key=lambda x: (x[0], x[1]))
            return {"mst_edges": mst_edges}
            
        except Exception as e:
            logging.error(f"Error in solve method: {e}")
            raise e
\end{lstlisting}
\begin{lstlisting}[language=Python, label={lst:mst_checkpoint_4}, caption=MST: code quality improvements., firstline=44, lastline=117]
import logging
from typing import Any, Dict, List, Optional

class MinimumSpanningTree:
    """
    Implementation of minimum_spanning_tree task using Kruskal's algorithm for improved performance.
    """
    
    def __init__(self):
        """Initialize the MinimumSpanningTree."""
        pass
    
    def solve(self, problem: Dict[str, Any]) -> Dict[str, List[List[float]]]:
        """
        Solve the minimum_spanning_tree problem using Kruskal's algorithm.
        
        Args:
            problem: Dictionary containing problem data specific to minimum_spanning_tree
                   
        Returns:
            The solution in the format expected by the task
        """
        try:
            num_nodes = problem["num_nodes"]
            edges = problem["edges"]

            # Sort edges by weight to facilitate Kruskal's algorithm
            edges_sorted = sorted(edges, key=lambda x: x[2])

            parent = list(range(num_nodes))
            size = [1] * num_nodes  # Using size for union by larger size

            def find(u: int) -> int:
                while parent[u] != u:
                    parent[u] = parent[parent[u]]  # Path compression
                    u = parent[u]
                return u
            
            def union(u: int, v: int) -> bool:
                u_root = find(u)
                v_root = find(v)
                if u_root == v_root:
                    return False  # Already connected
                # Union by size: attach smaller to larger
                if size[u_root] < size[v_root]:
                    parent[u_root] = v_root
                    size[v_root] += size[u_root]
                else:
                    parent[v_root] = u_root
                    size[u_root] += size[v_root]
                return True

            mst_edges = []
            edge_count = 0

            for edge in edges_sorted:
                u, v, w = edge
                if find(u) != find(v):
                    union(u, v)
                    # Ensure consistent ordering for output sorting
                    if u > v:
                        u, v = v, u
                    mst_edges.append([u, v, w])
                    edge_count += 1
                    if edge_count == num_nodes - 1:
                        break  # Early exit once MST is complete

            # Final sort by (u, v) for consistent output
            mst_edges.sort(key=lambda x: (x[0], x[1]))
            return {"mst_edges": mst_edges}
            
        except Exception as e:
            logging.error(f"Error in solve method: {e}")
            raise e
\end{lstlisting}
\begin{lstlisting}[language=Python, label={lst:mst_checkpoint_12}, caption=MST: micro-optimized implementation., firstline=44, lastline=107]
import logging
import operator
from typing import Any, Dict, List, Optional

class MinimumSpanningTree:
    """
    Implementation of minimum_spanning_tree task using Kruskal's algorithm for improved performance.
    """
    
    def __init__(self):
        pass  # No initialization needed
    
    def solve(self, problem: Dict[str, Any]) -> Dict[str, List[List[float]]]:
        try:
            num_nodes = problem["num_nodes"]
            edges = problem["edges"]

            # Sort edges by weight using efficient operator.itemgetter
            edges_sorted = sorted(edges, key=operator.itemgetter(2))

            parent = list(range(num_nodes))
            size = [1] * num_nodes

            def find(u: int) -> int:
                while parent[u] != u:
                    parent[u] = parent[parent[u]]  # Path compression
                    u = parent[u]
                return u
            
            def union(u: int, v: int) -> bool:
                u_root = find(u)
                v_root = find(v)
                if u_root == v_root:
                    return False
                if size[u_root] < size[v_root]:
                    parent[u_root] = v_root
                    size[v_root] += size[u_root]
                else:
                    parent[v_root] = u_root
                    size[u_root] += size[v_root]
                return True

            mst_edges = []
            edge_count = 0
            target = num_nodes - 1  # Precompute to avoid repeated calculation

            for edge in edges_sorted:
                u, v, w = edge
                if find(u) != find(v):
                    union(u, v)
                    if u > v:
                        u, v = v, u
                    mst_edges.append([u, v, w])
                    edge_count += 1
                    if edge_count == target:
                        break

            # Final sort using operator for efficiency
            mst_edges.sort(key=operator.itemgetter(0, 1))
            return {"mst_edges": mst_edges}
            
        except Exception as e:
            logging.error(f"Error in solve method: {e}")
            raise e
\end{lstlisting}

\subsection{Circle Packing}
\label{app:openevolve_circle_packing}
\begin{figure}[!t]
    \centering
    \subfloat[SemRep]{
        \includegraphics[width=0.22\linewidth]{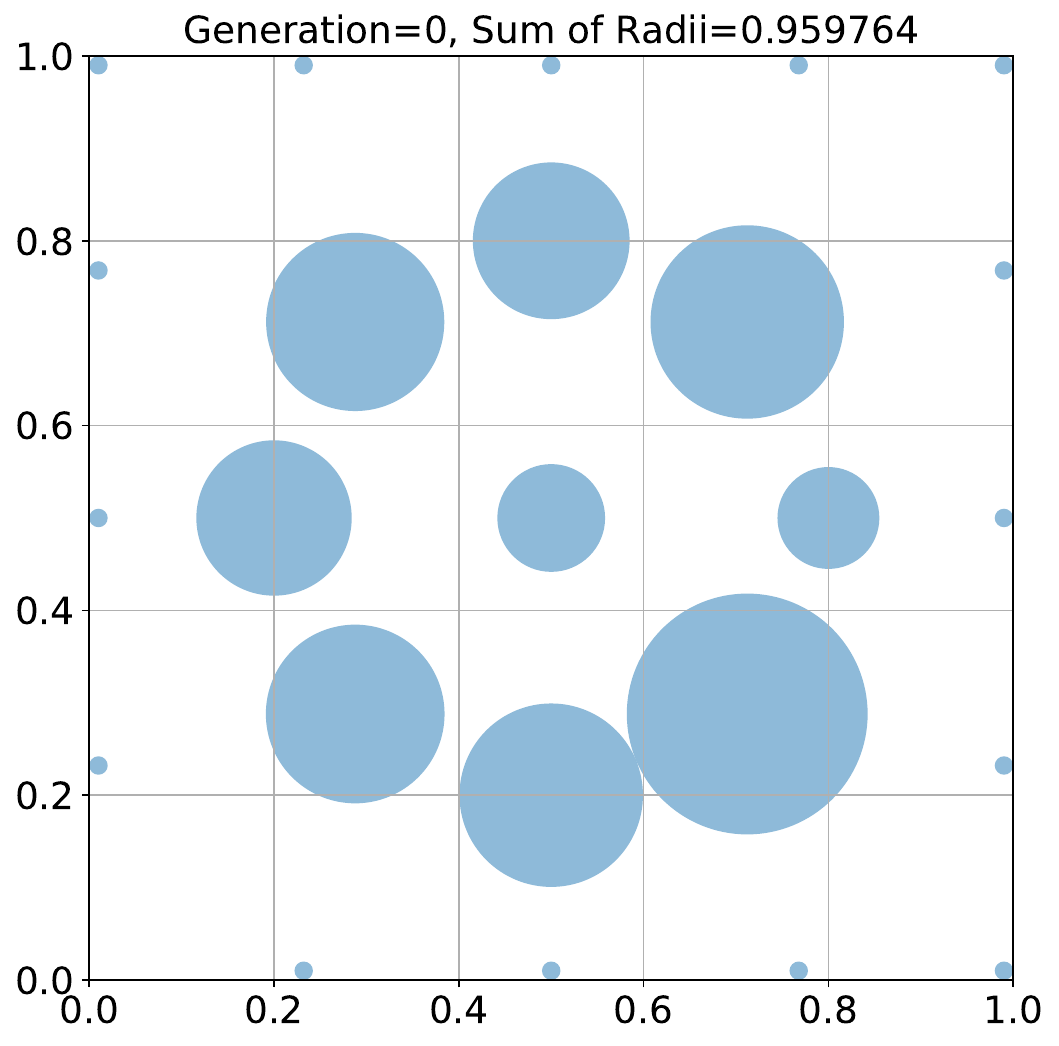}
        \includegraphics[width=0.22\linewidth]{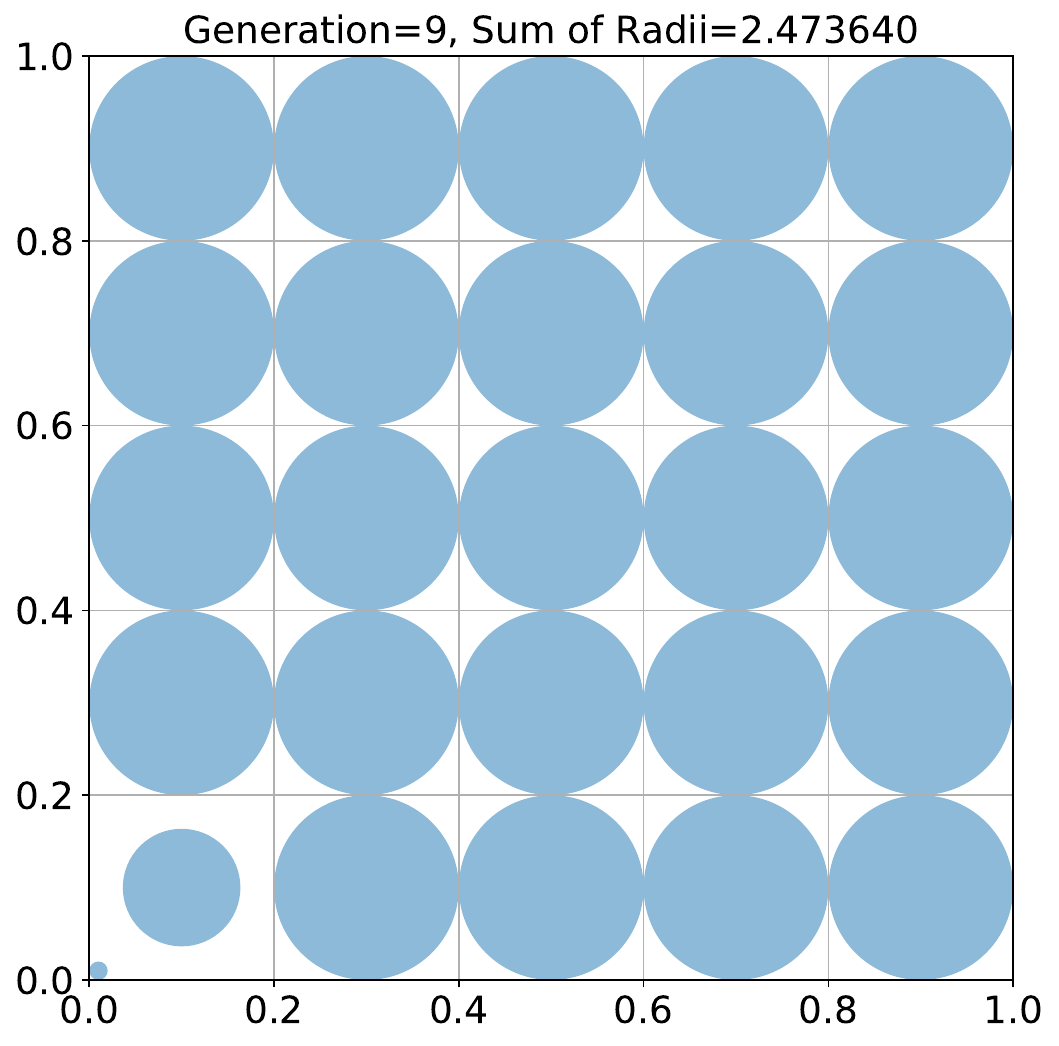}
        \includegraphics[width=0.22\linewidth]{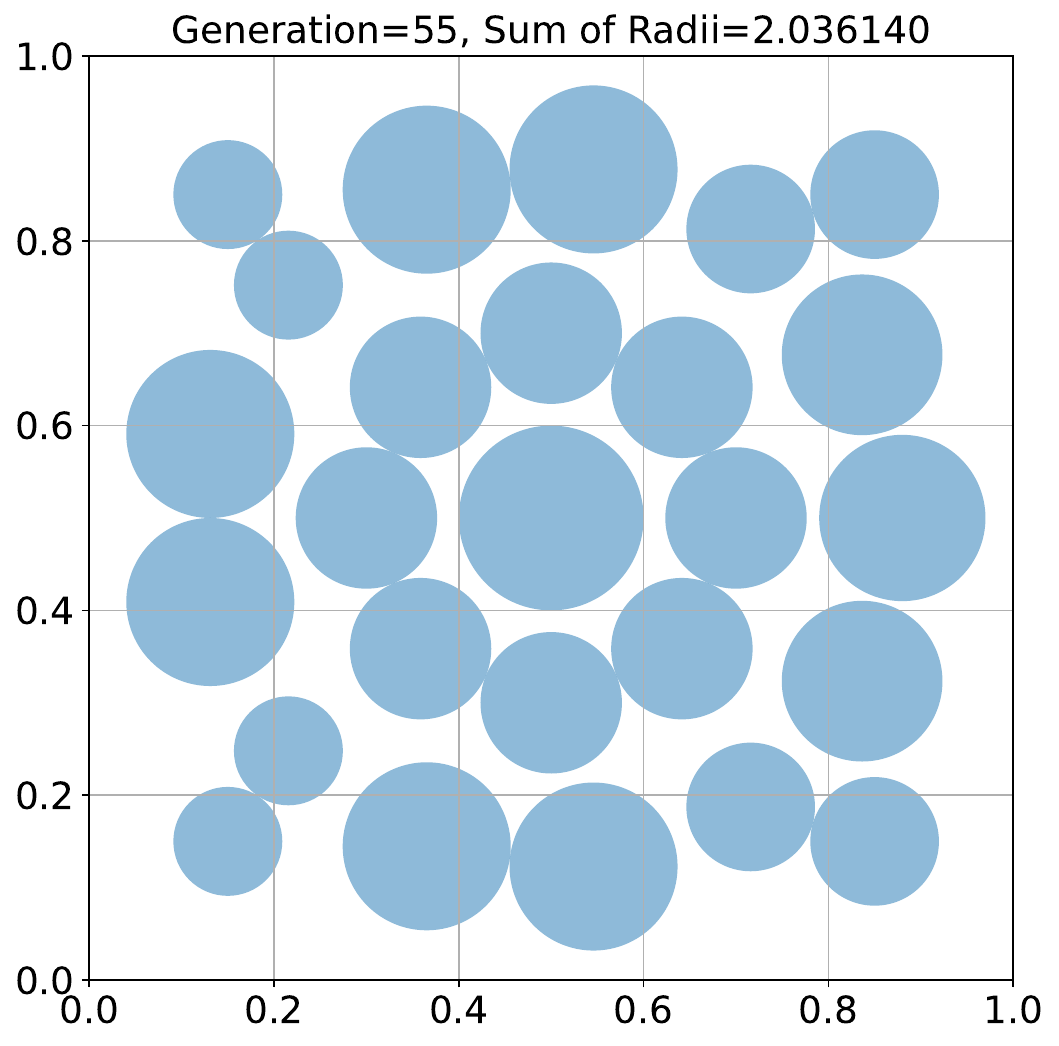}
        \includegraphics[width=0.22\linewidth]{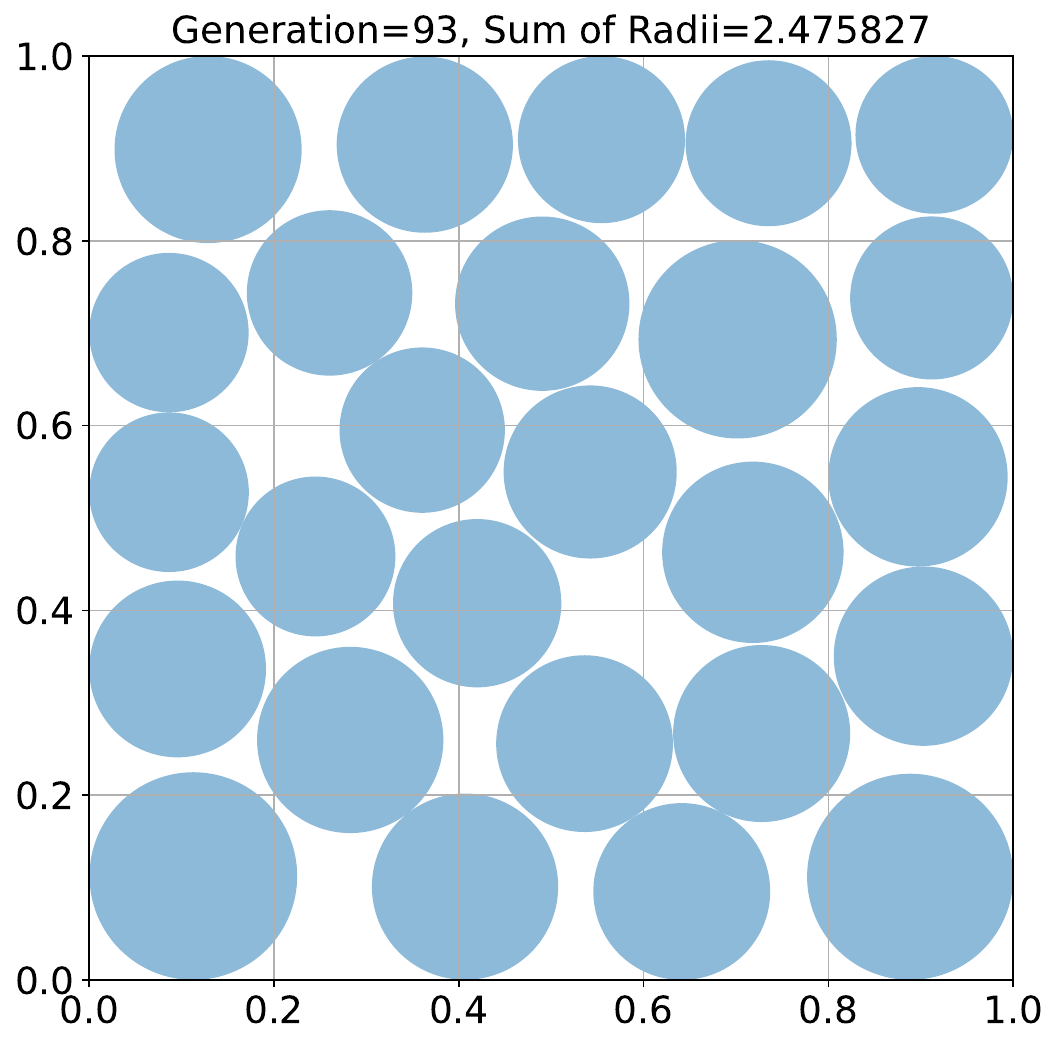}
    }

    \subfloat[QwQ-32B]{
        \includegraphics[width=0.22\linewidth]{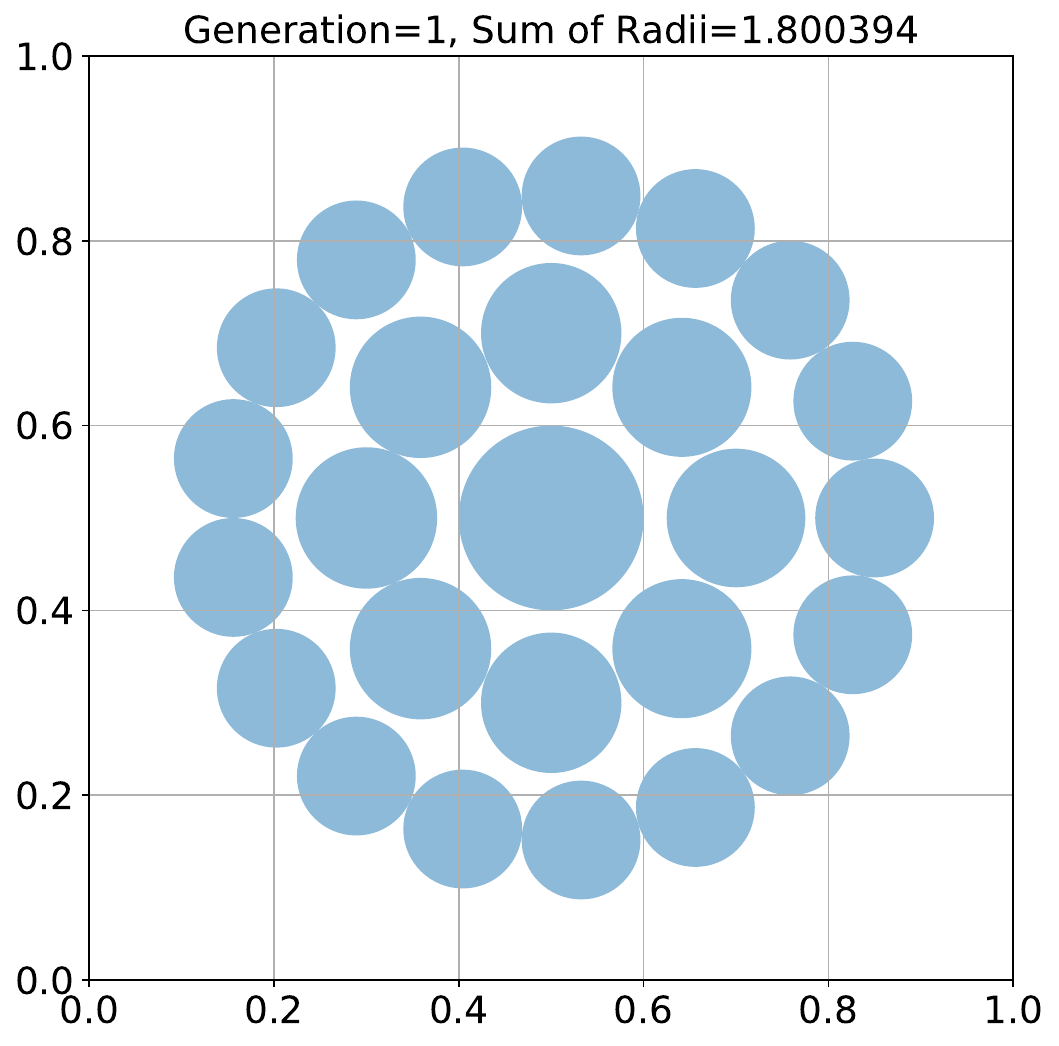}
        \includegraphics[width=0.22\linewidth]{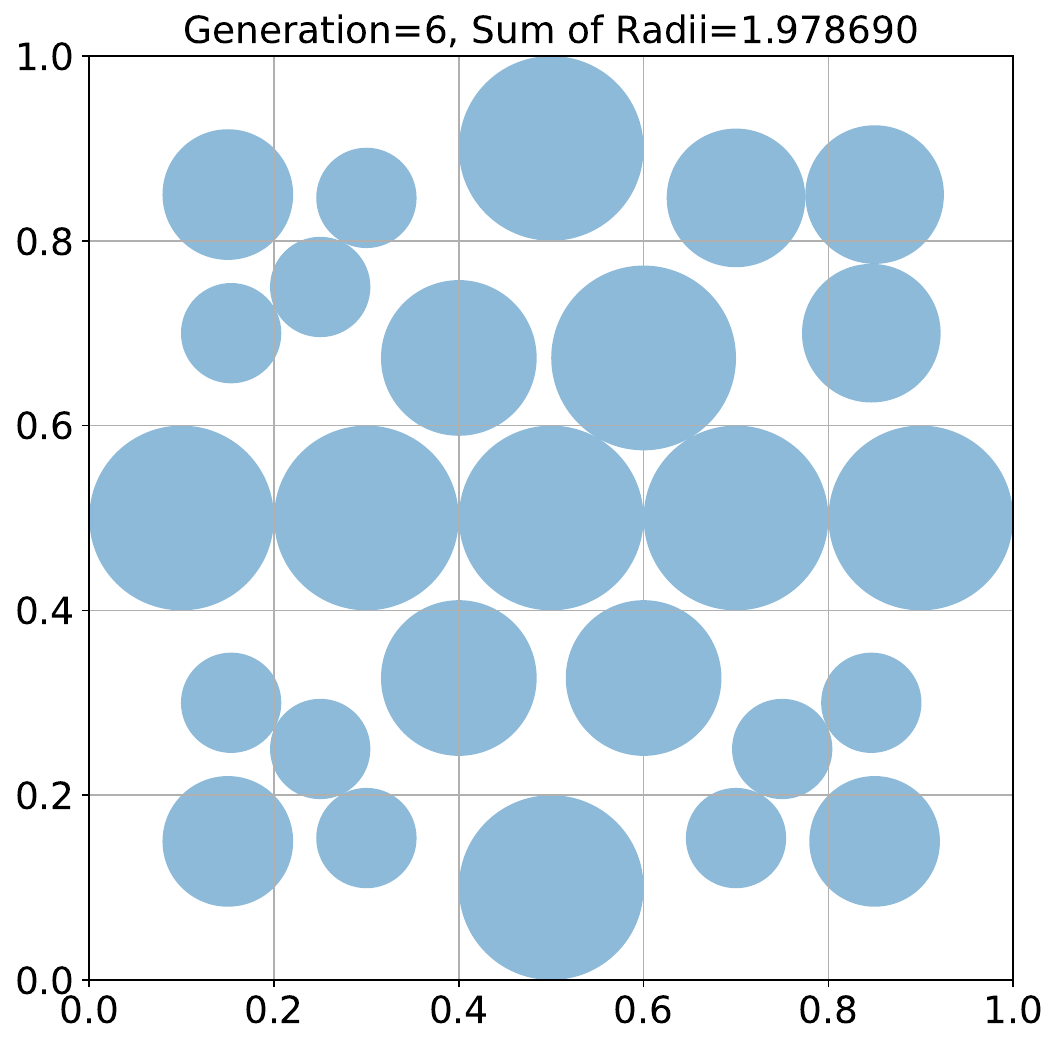}
        \includegraphics[width=0.22\linewidth]{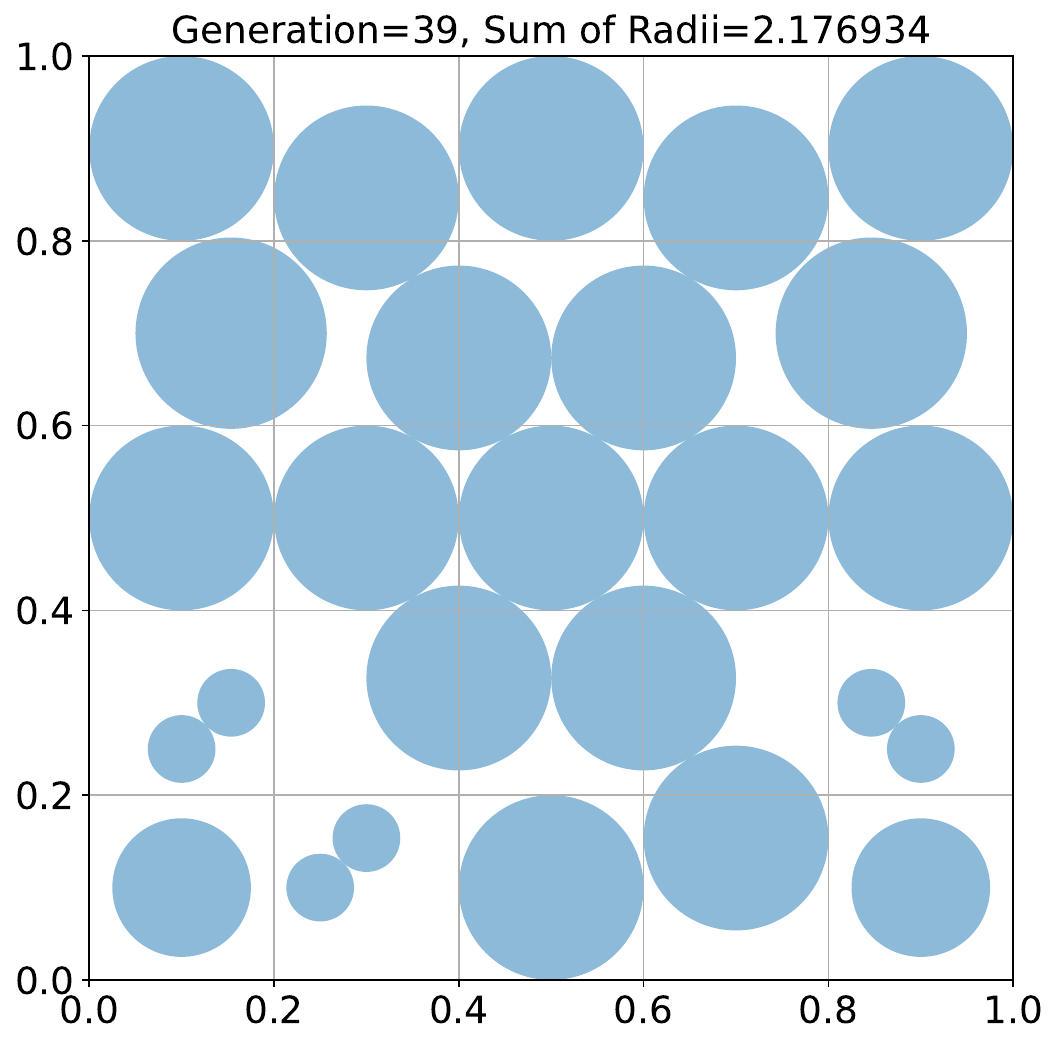}
        \includegraphics[width=0.22\linewidth]{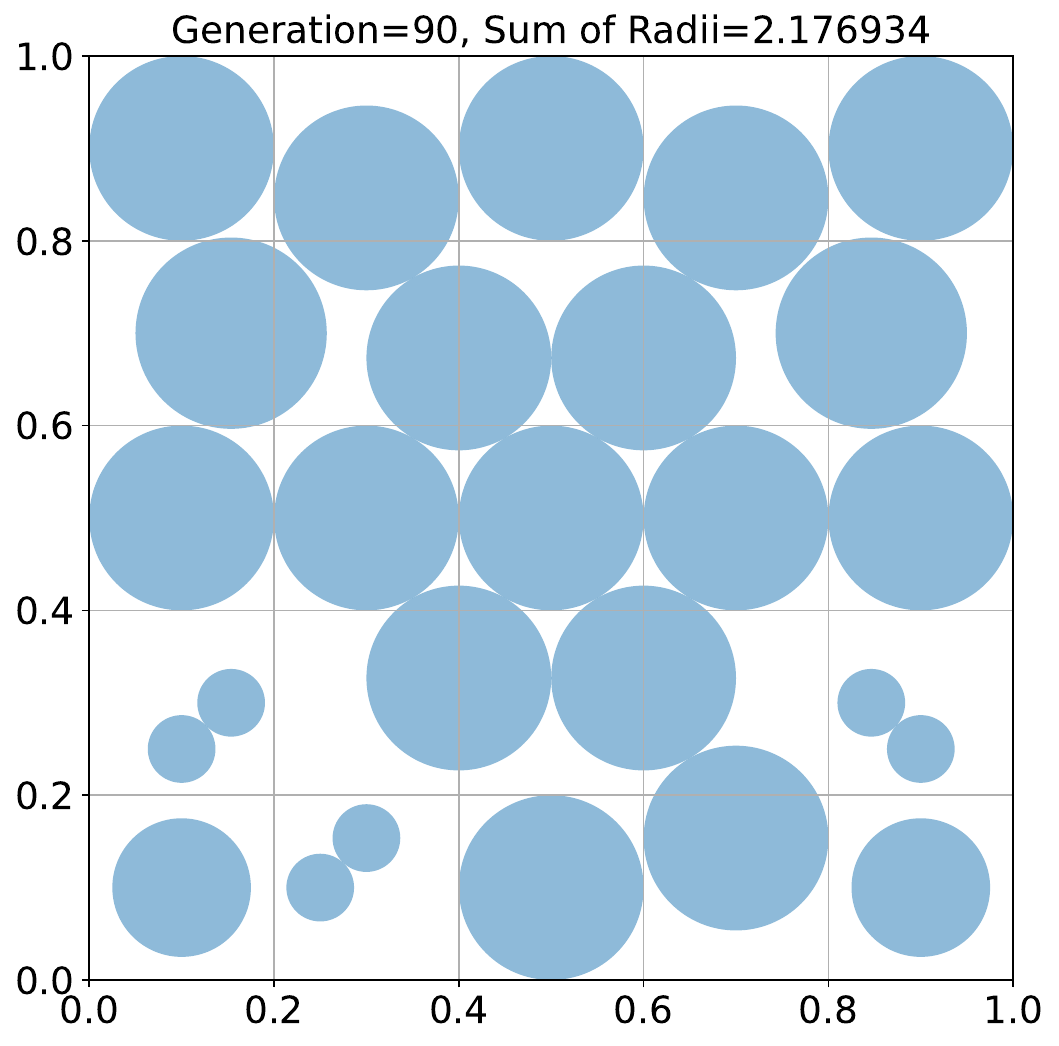}
    }

    \subfloat[DeepSeek-V3-Chat]{
        \includegraphics[width=0.22\linewidth]{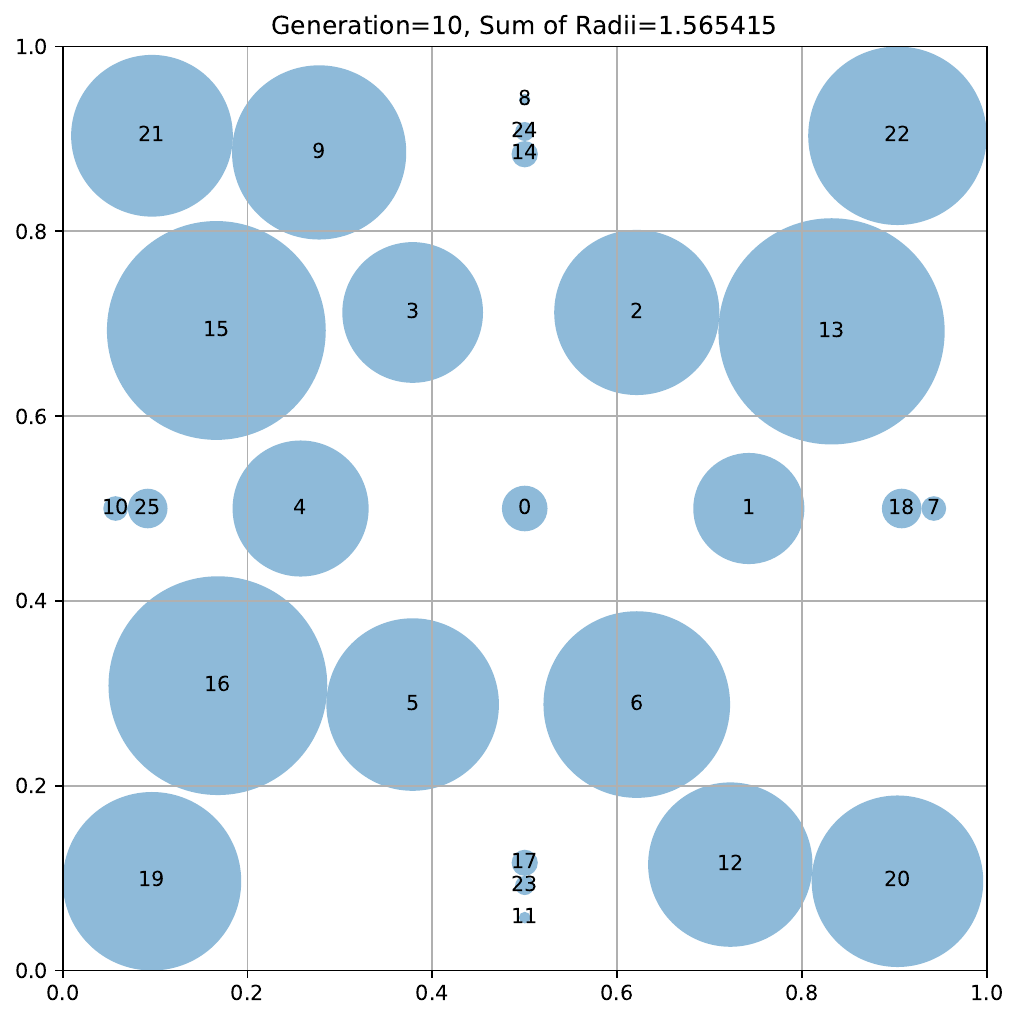}
        \includegraphics[width=0.22\linewidth]{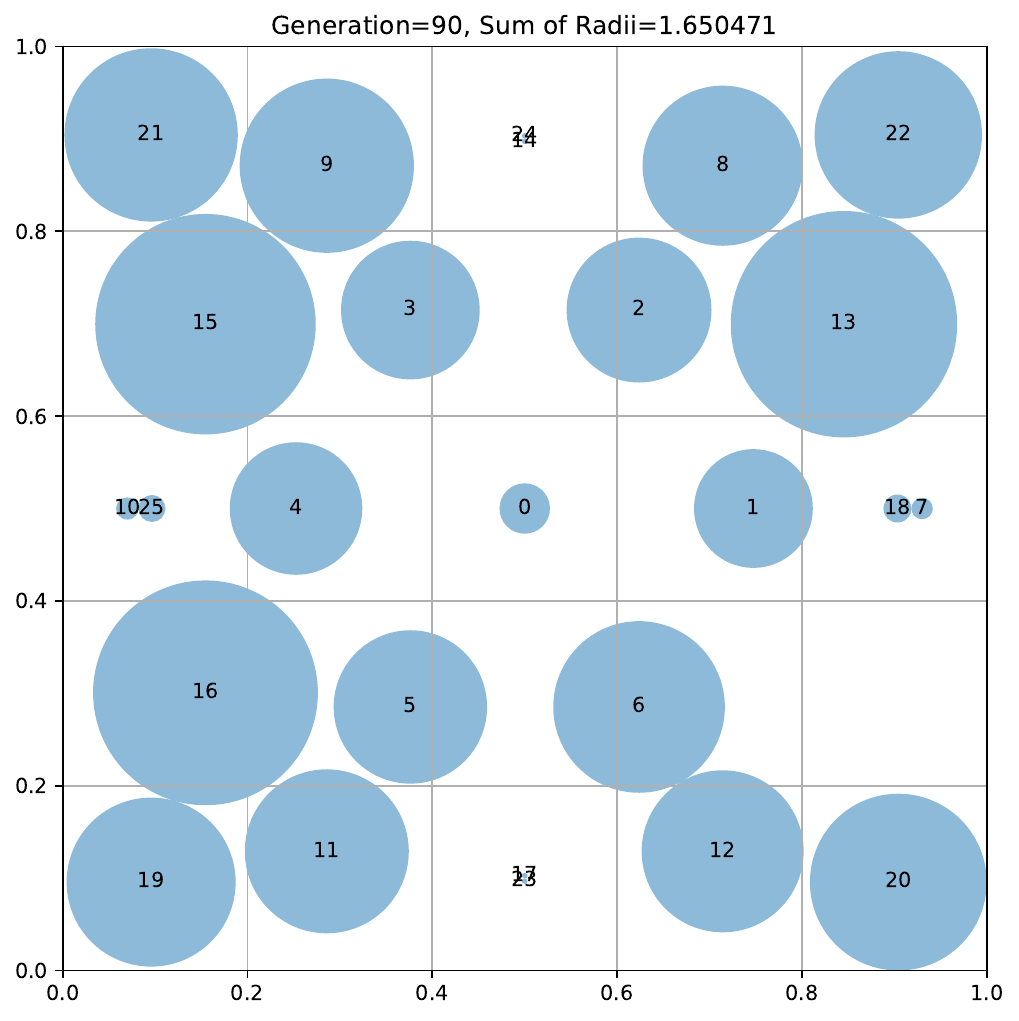}
        \includegraphics[width=0.22\linewidth]{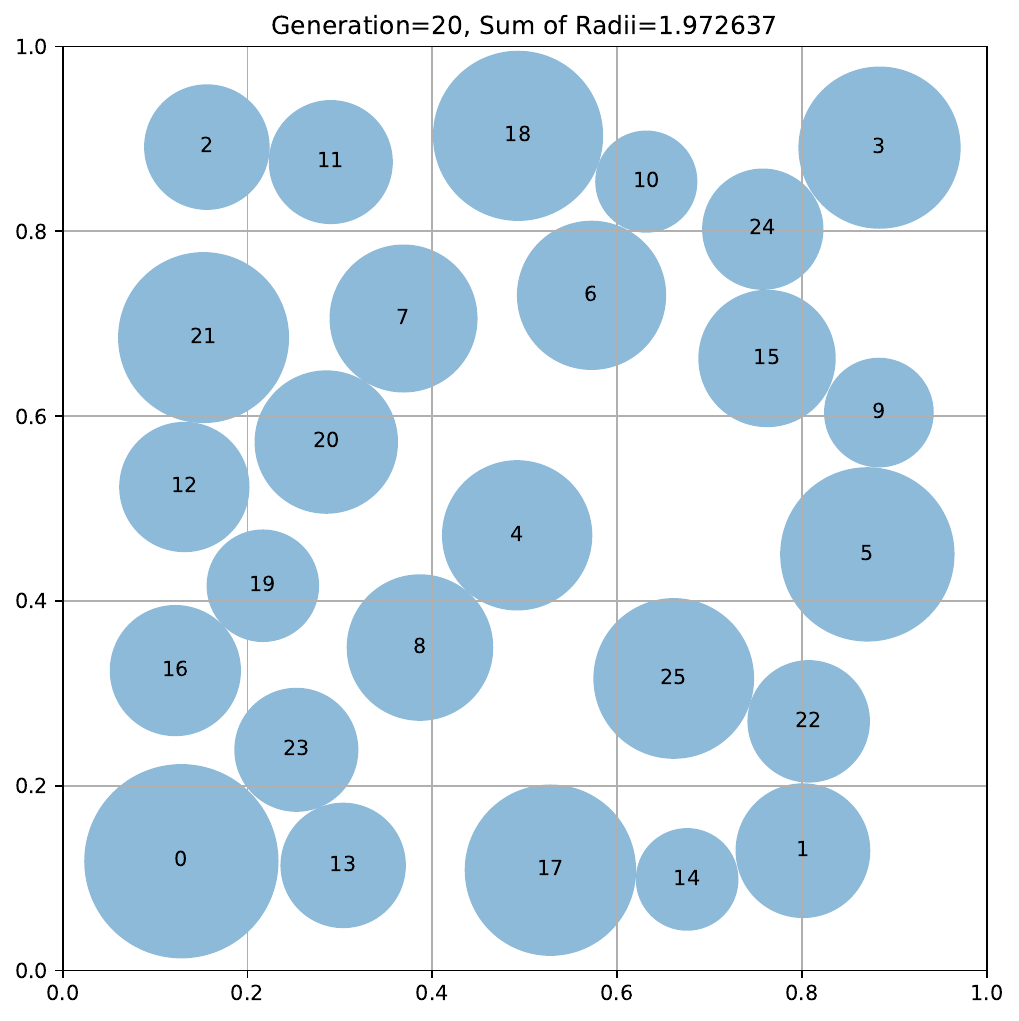}
        \includegraphics[width=0.22\linewidth]{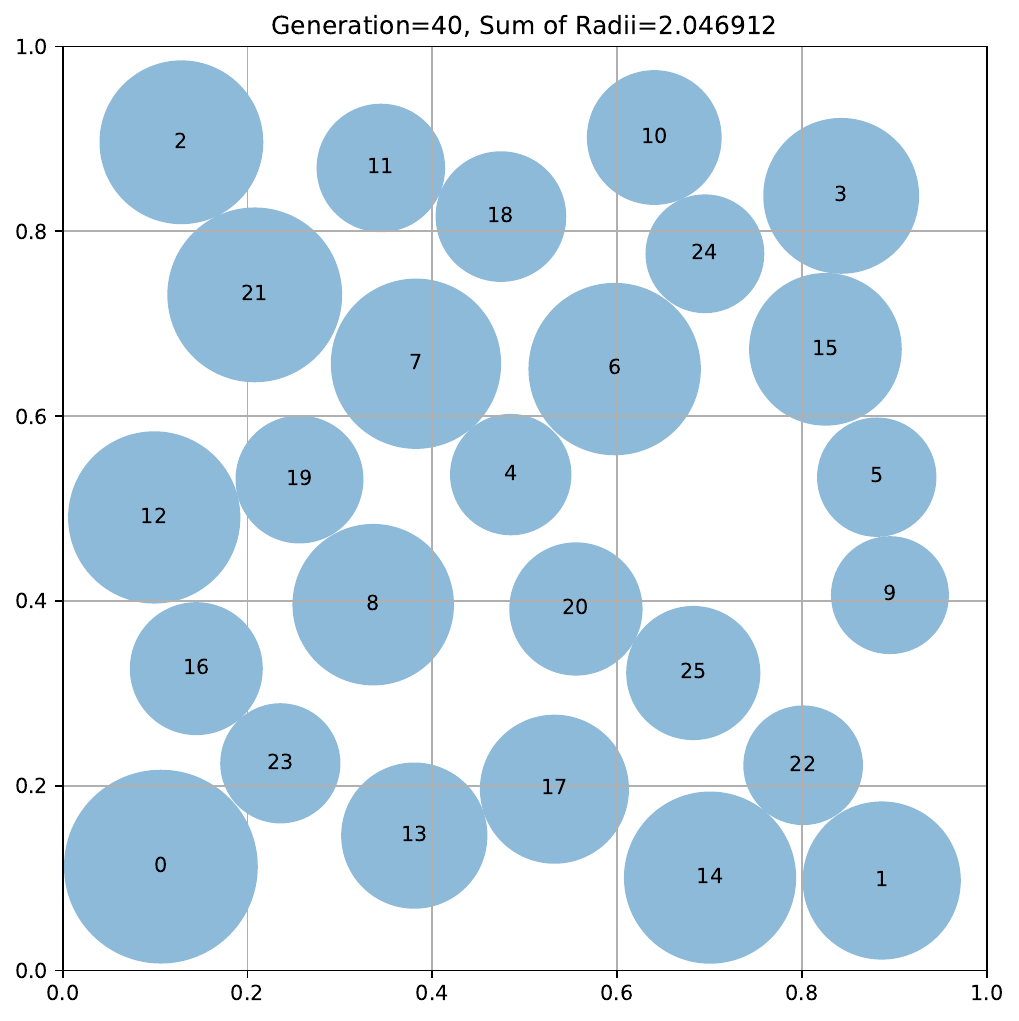}
    }
    
    \subfloat[DeepSeek-V3-Reasoner]{
        \includegraphics[width=0.22\linewidth]{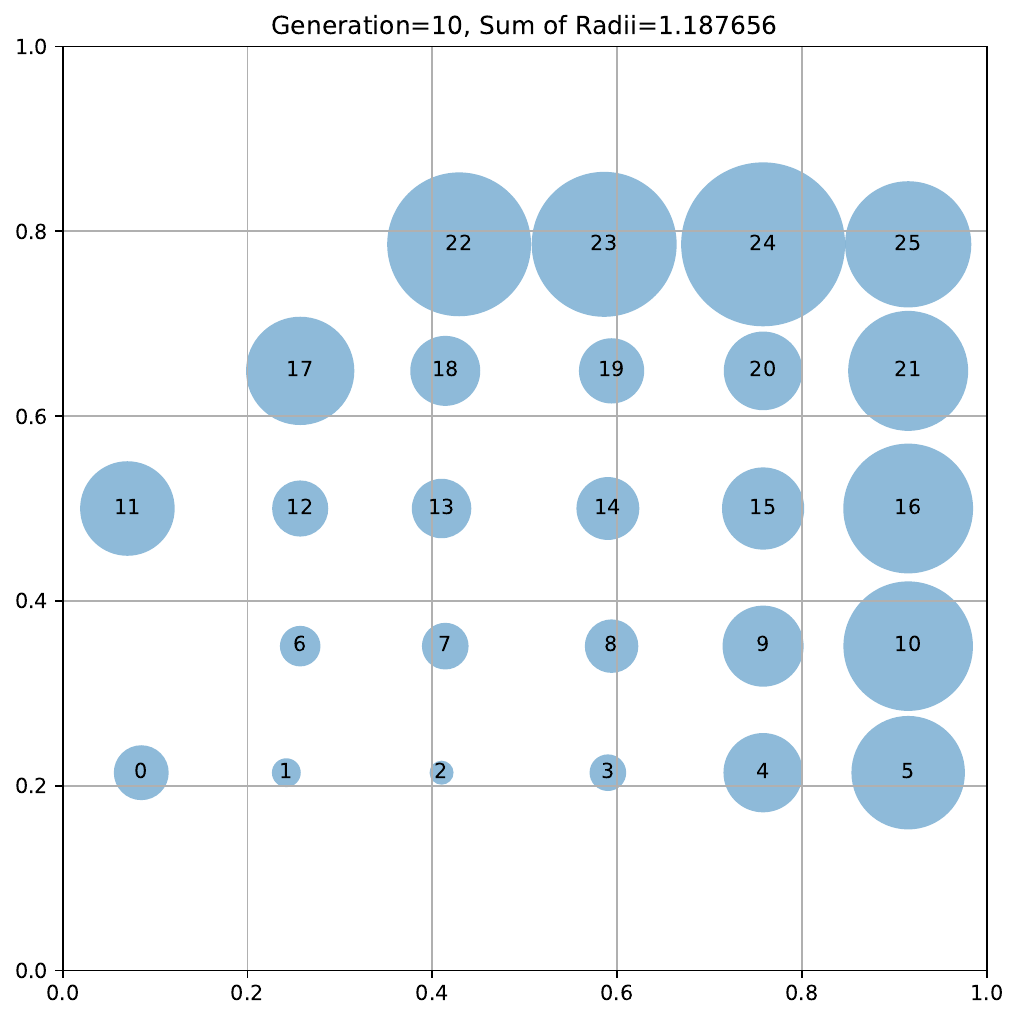}
        \includegraphics[width=0.22\linewidth]{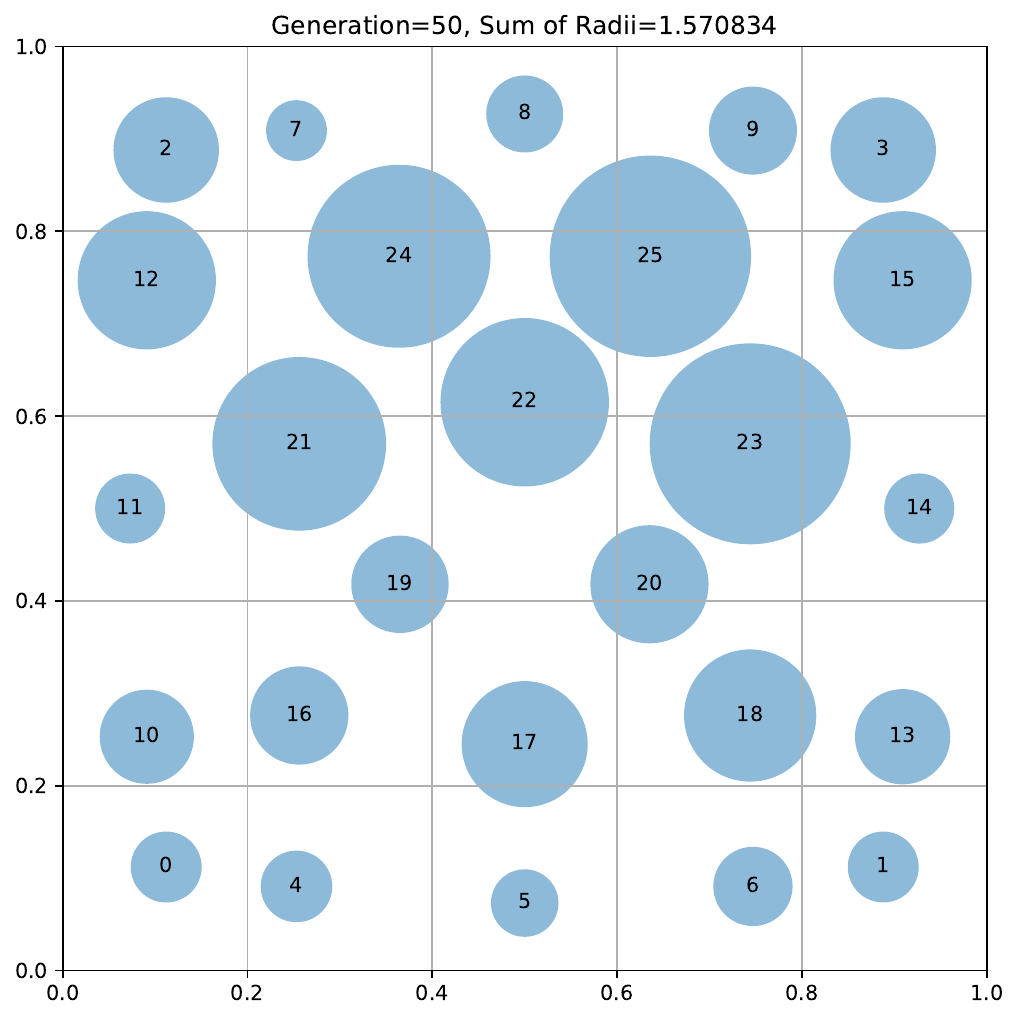}
        \includegraphics[width=0.22\linewidth]{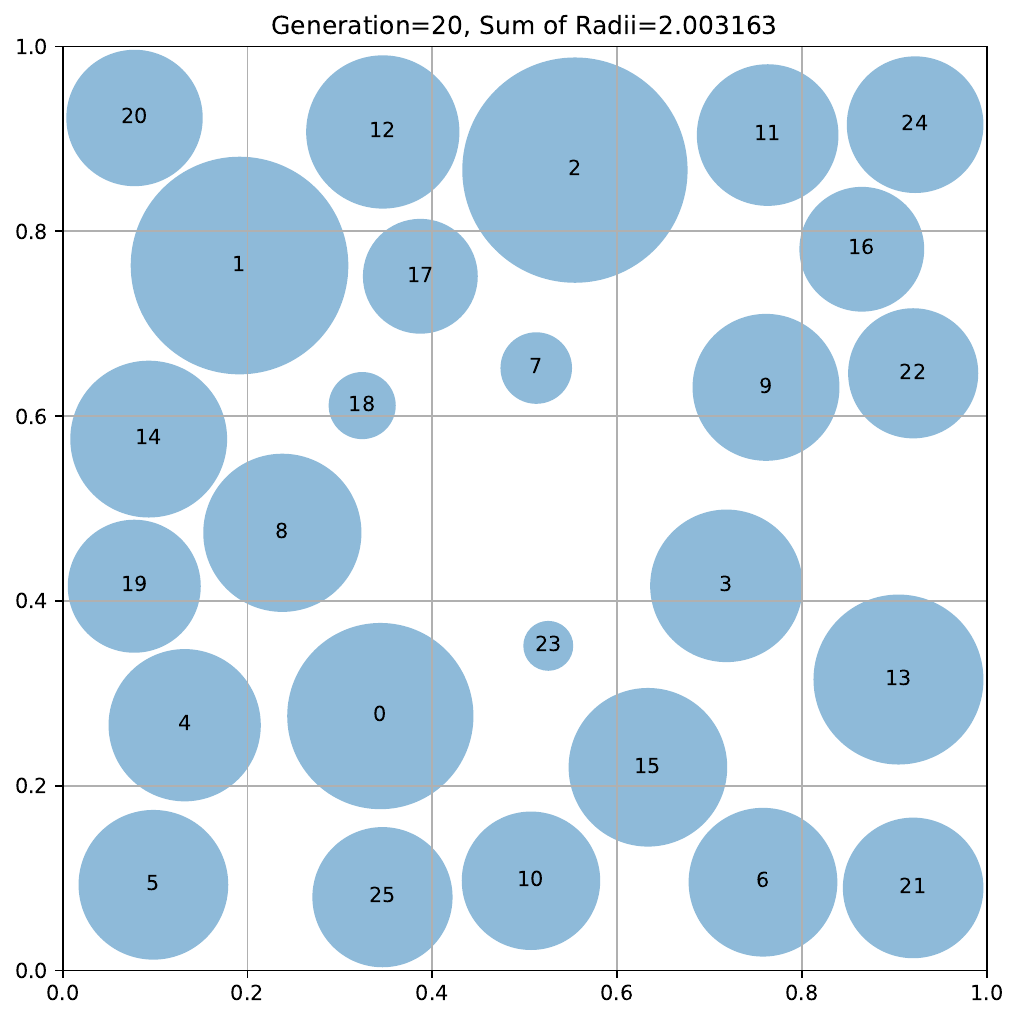}
        \includegraphics[width=0.22\linewidth]{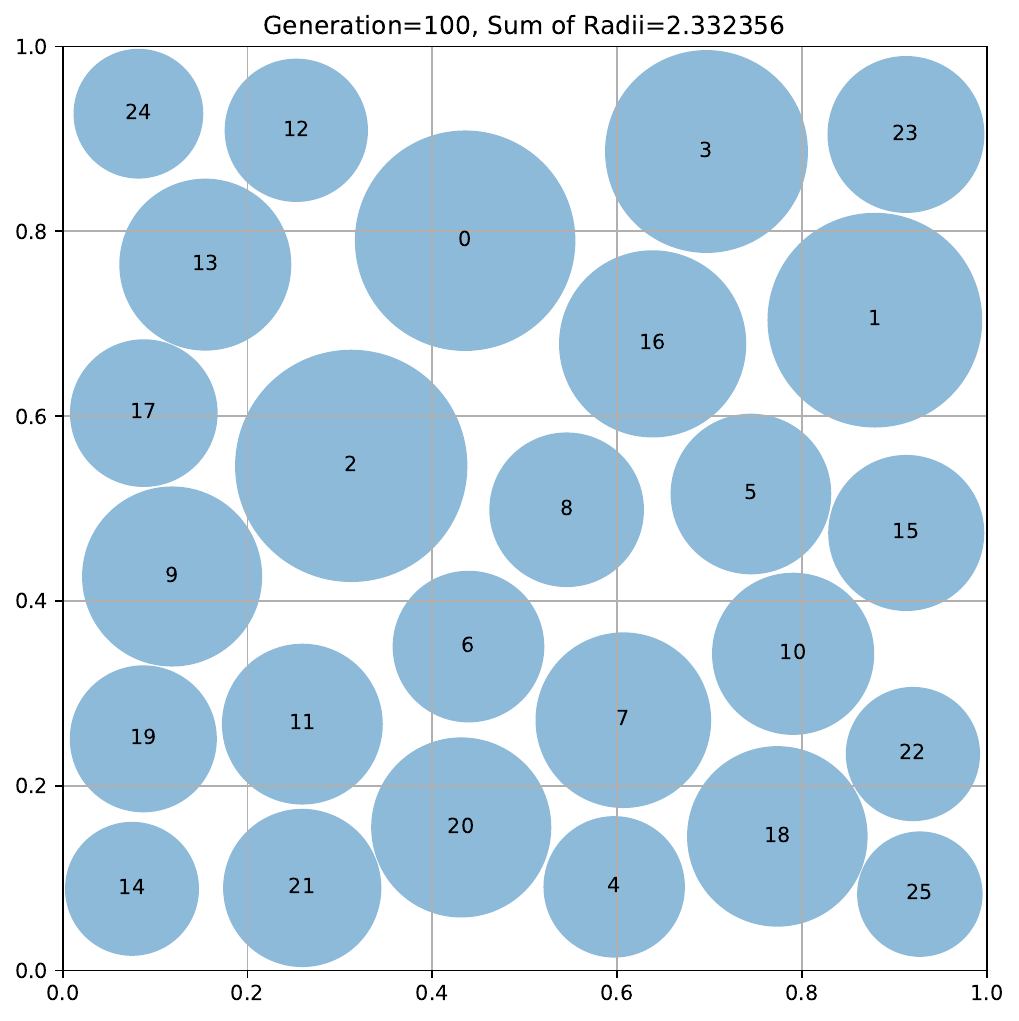}
    }
    \caption{Visualized trajectories of circle packing optimization, illustrating the iterative refinement of configurations across different models.}
    \label{fig:circle_packing_trajs}
\end{figure}
\paragraph{Hyperparameters.}
We employ a two-phase evolutionary search starting from our baseline. The LLM configuration is the same as AlgoTune. 
Phase 1 uses a population size of 60, an archive size of 25, 4 islands, an elite selection ratio $0.3$, an exploitation ratio $0.7$, the top 3 programs in prompts, and runs for 100 iterations. 
Phase 2 uses a population size of 70, an archive size of 30, 5 islands, an elite selection ratio $0.3$, an exploitation ratio $0.6$, the top 4 programs in prompts, and runs for 100 iterations.

\paragraph{Trajectory analysis.}
The state-of-the-art result for this circle packing problem is 2.6358, achieved by AlphaEvolve~\cite{novikov2025alphaevolve}.
We track the evolution trajectories of \sys in detail, examining important breakthroughs at each stage of its evolution:
\begin{itemize}
    \item \textbf{Early discovery (\cref{lst:circle_packing_checkpoint_9}).} 
    \sys discovers an effective grid-based pattern achieving 2.4736 (93.88\% of the target) at iteration 9.
    Specifically, \sys uses a systematic grid placement with the 26th circle at the true center (0.5, 0.5) and improved radius computation using edge distances and half the closest neighbor distance.
    \item \textbf{Exploration (\cref{lst:circle_packing_checkpoint_55}).}
    \sys abandons the grid pattern and explores a ring-based approach, which reduces the performance to 2.036 (77.27\% of the target).
    This demonstrates that \sys actively explores diverse semantic-preserving representations that might even lead to temporary performance degradation.
    \item \textbf{Final optimization (\cref{lst:circle_packing_checkpoint_93}).}
    Building on the ring-based representation discovered during exploration, \sys achieves the highest score of 2.4737 through extensive simulated annealing, surpassing the previous peak.
    This final outcome is consistent with our intuition that a temporary degradation in performance might expose additional optimization opportunities and ultimately lead to superior outcomes.
\end{itemize}

We also illustrate the evolution trajectories for all models in \cref{fig:circle_packing_trajs}, offering a high-level view of how each model progressively refines circle arrangements through iterative optimization.

\begin{lstlisting}[language=Python, label={lst:circle_packing_checkpoint_9}, caption=Circle packing implementation at iteration 9: grid-based pattern with improved radius computation., firstline=2, lastline=78]
"""Constructor-based circle packing for n=26 circles"""
import numpy as np

def construct_packing():
    """
    Construct a specific arrangement of 26 circles in a unit square
    that attempts to maximize the sum of their radii.

    Returns:
        Tuple of (centers, radii, sum_of_radii)
        centers: np.array of shape (26, 2) with (x, y) coordinates
        radii: np.array of shape (26) with radius of each circle
        sum_of_radii: Sum of all radii
    """
    n = 26
    centers = np.zeros((n, 2))

    # Create a 5x5 grid with spacing 0.2, skipping the central cell
    spacing = 0.2
    idx = 0
    for i in range(5):
        y = 0.1 + i * spacing
        for j in range(5):
            x = 0.1 + j * spacing
            if i == 2 and j == 2:  # Skip the center cell (0.5, 0.5)
                continue
            centers[idx] = [x, y]
            idx += 1

    # Add the 26th circle at the true center
    centers[25] = [0.5, 0.5]

    # Clip to ensure all circles are within the unit square
    centers = np.clip(centers, 0.01, 0.99)

    # Compute radii using the improved method
    radii = compute_max_radii(centers)
    sum_radii = np.sum(radii)

    return centers, radii, sum_radii

def compute_max_radii(centers):
    """
    Compute the maximum valid radii for each circle position without overlapping or exceeding edges.
    Uses edge distances and half the closest neighbor distance.

    Args:
        centers: np.array of shape (n, 2) with (x, y) coordinates

    Returns:
        np.array of shape (n) with radius of each circle
    """
    n = centers.shape[0]
    radii = np.empty(n)

    # Compute edge distances for each circle
    edge_dists = np.minimum(centers[:, 0], 1 - centers[:, 0])
    edge_dists = np.minimum(edge_dists, centers[:, 1])
    edge_dists = np.minimum(edge_dists, 1 - centers[:, 1])

    # Compute minimum distance to any other circle divided by 2
    min_dists = np.full(n, np.inf)
    for i in range(n):
        for j in range(i + 1, n):
            dist = np.linalg.norm(centers[i] - centers[j])
            if dist < min_dists[i]:
                min_dists[i] = dist
            if dist < min_dists[j]:
                min_dists[j] = dist
    min_dists /= 2.0  # Max possible radius without overlapping closest neighbor

    # The radius is the minimum of edge distance and min neighbor distance/2
    radii = np.minimum(edge_dists, min_dists)

    return radii

# EVOLVE-BLOCK-END
\end{lstlisting}
\begin{lstlisting}[language=Python, label={lst:circle_packing_checkpoint_55}, caption=Circle packing implementation at iteration 55: exploratory ring-based pattern., firstline=2, lastline=57]
import numpy as np

def construct_packing():
    n = 26
    centers = np.zeros((n, 2))

    # Central circle
    centers[0] = [0.5, 0.5]

    # First ring (8 circles) at radius 0.2 from center
    for i in range(8):
        angle = 2 * np.pi * i / 8
        centers[i + 1] = [0.5 + 0.2 * np.cos(angle), 0.5 + 0.2 * np.sin(angle)]

    # Outer ring (13 circles) at radius 0.38 from center
    for i in range(13):
        angle = 2 * np.pi * i / 13
        centers[9 + i] = [0.5 + 0.38 * np.cos(angle), 0.5 + 0.38 * np.sin(angle)]

    # Add four corner circles
    corners = [
        [0.15, 0.15],  # Bottom-left corner
        [0.85, 0.15],  # Bottom-right corner
        [0.85, 0.85],  # Top-right corner
        [0.15, 0.85]   # Top-left corner
    ]
    centers[22:26] = corners

    radii = compute_max_radii(centers)
    sum_radii = np.sum(radii)

    return centers, radii, sum_radii

def compute_max_radii(centers):
    n = centers.shape[0]
    radii = np.zeros(n)

    # Edge distances
    edge_dists = np.minimum(centers[:, 0], centers[:, 1])
    edge_dists = np.minimum(edge_dists, np.minimum(1 - centers[:, 0], 1 - centers[:, 1]))

    # Nearest neighbor distances
    min_neighbor_dists = np.full(n, np.inf)
    for i in range(n):
        for j in range(i + 1, n):
            dist = np.linalg.norm(centers[i] - centers[j])
            if dist < min_neighbor_dists[i]:
                min_neighbor_dists[i] = dist
            if dist < min_neighbor_dists[j]:
                min_neighbor_dists[j] = dist

    # Compute radii as the minimum between edge and half neighbor distance
    radii = np.minimum(edge_dists, min_neighbor_dists / 2)

    return radii
\end{lstlisting}
\begin{lstlisting}[language=Python, label={lst:circle_packing_checkpoint_93}, caption=Circle packing implementation at iteration 93: ring-based initialization with extensive simulated annealing., firstline=2, lastline=85]
import numpy as np

def construct_packing():
    n = 26
    centers = np.zeros((n, 2))

    # Central circle
    centers[0] = [0.5, 0.5]

    # First ring (8 circles) at radius 0.2 from center
    for i in range(8):
        angle = 2 * np.pi * i / 8
        centers[i + 1] = [
            0.5 + 0.2 * np.cos(angle),
            0.5 + 0.2 * np.sin(angle)
        ]

    # Outer ring (13 circles) at radius 0.35 from center (adjusted radius)
    for i in range(13):
        angle = 2 * np.pi * i / 13
        centers[9 + i] = [
            0.5 + 0.35 * np.cos(angle),
            0.5 + 0.35 * np.sin(angle)
        ]

    # Add four corner circles placed closer to the edges (adjusted positions)
    corners = [
        [0.1, 0.1],  # Bottom-left corner
        [0.9, 0.1],  # Bottom-right corner
        [0.9, 0.9],  # Top-right corner
        [0.1, 0.9]   # Top-left corner
    ]
    centers[22:26] = corners

    # Simulated annealing optimization
    best_centers = centers.copy()
    best_sum = compute_sum(best_centers)
    current_centers = best_centers.copy()
    current_sum = best_sum
    temp = 1.0
    cooling = 0.9995
    steps = 200000  # Increased steps for better optimization

    for _ in range(steps):
        idx = np.random.randint(n)
        delta = np.random.uniform(-0.015, 0.015, 2)  # Slightly larger delta for exploration
        new_center = current_centers[idx] + delta
        new_center = np.clip(new_center, 0.0, 1.0)
        new_centers = current_centers.copy()
        new_centers[idx] = new_center
        new_sum = compute_sum(new_centers)

        # Metropolis criterion with improved acceptance probability
        if new_sum > current_sum or np.random.rand() < np.exp((new_sum - current_sum) / temp):
            current_centers = new_centers.copy()
            current_sum = new_sum
            if new_sum > best_sum:
                best_centers = new_centers.copy()
                best_sum = new_sum
        temp *= cooling

    return best_centers, compute_max_radii(best_centers), best_sum

def compute_max_radii(centers):
    n = centers.shape[0]
    radii = np.zeros(n)

    # Edge distances to all four sides
    edge_dists = np.minimum(centers[:, 0], centers[:, 1])
    edge_dists = np.minimum(edge_dists, np.minimum(1 - centers[:, 0], 1 - centers[:, 1]))

    # Nearest neighbor distances using efficient computation
    dist_matrix = np.linalg.norm(centers[:, np.newaxis] - centers, axis=2)
    np.fill_diagonal(dist_matrix, np.inf)
    min_neighbor_dists = np.min(dist_matrix, axis=1)

    # Radii are the minimum between edge distance and half the nearest neighbor distance
    radii = np.minimum(edge_dists, min_neighbor_dists / 2)

    return radii

def compute_sum(centers):
    return np.sum(compute_max_radii(centers))
\end{lstlisting}

\section{EditBench Case Study}
\label{app:editbench_case_study}
We show the full code of the case study discussed in \cref{subsec:case_study}.
\cref{lst:editbench_input} is the given input.
\cref{lst:editbench_ours_step_1_diff} and \cref{lst:editbench_ours_step_2_diff} demonstrate how \sys takes advantage of the semantics-preserving transformation and then performs the order-consistent parallelism by using \texttt{asyncio.gather}.
\cref{lst:editbench_direct_step_1_diff} and \cref{lst:editbench_direct_step_2_diff} show how existing methods fail to preserve output ordering while successfully performing the parallelism optimization by using \texttt{asyncio.as\_completed}.

\begin{lstlisting}[language=Python, label={lst:editbench_input}, caption=EditBench input code.]
import logging
import os
from typing import Any, Dict, List
from pydantic import BaseModel, Field
from carvana_enzo_worker.enums.gpt_enums import GptModels, VertextAIModels
from carvana_enzo_worker.providers.vertexai_claude_provider import VertexAIClaudeProvider
from carvana_enzo_worker.providers.vertexai_gemini_provider import VertexAIGeminiProvider
from carvana_enzo_worker.providers.azure_o1_provider import AzureOpenAIo1Provider
from carvana_enzo_worker.providers.azure_gpt_provider import AzureOpenAIChatProvider

# pylint: disable=W1203, C0415 [Use %s formatting in logging function, import-outside-toplevel]


class LLMArena(BaseModel):
    """
    A tool to generate chats using multiple LLM's for a given prompt
    """

    prompt: str = Field(..., description="The input prompt for the LLMs.")
    models: List[str] = Field(..., description="A list of model names to use for generating chats.")
    responses: List[str] = Field([], description="A list of generated chat responses.")
    kwargs: Dict[str, Any] = Field({}, description="Additional keyword arguments for the LLMs.")


    @staticmethod
    async def generate_responses_for_models(prompt: str, models: List[str], **kwargs: Any) -> List[str]:
        """
        Generate responses from multiple models for a given prompt.

        :param prompt: The input prompt for the LLMs.
        :param models: A list of model names to use for generating responses.
        :return: A list of generated responses.
        """
        responses = []
        providers = []
        for model in models:
            provider_for_model = LLMArena._get_provider_for_model(model, **kwargs)
            providers.append(provider_for_model)

        
        for provider in providers:
            try:
                response = await provider.generate_chat_response(prompt)
                responses.append(response)
            except Exception as e:
                logging.error(f"Error generating response from {provider}: {e}")
                responses.append(f"Error generating response from {provider}: {e}")

        return responses
    

    @staticmethod
    def _get_provider_for_model(model: str, **kwargs: Any) -> Any:
        event_id = kwargs.get("event_id", "")

        if model == VertextAIModels.CLAUDE_3_5_SONNET_V2.name:
            return  VertexAIClaudeProvider(event_id=event_id, location=str(os.getenv("VERTEXAI_CLAUDE_REGION")), deployment_id=model)
        
        if model == VertextAIModels.GEMINI_2_0_FLASH_EXP.name:
            return VertexAIGeminiProvider(event_id=event_id, location=str(os.getenv("VERTEXAI_GEMINI_REGION")), deployment_id=model)
        
        if model == GptModels.o1.value:
            return AzureOpenAIo1Provider(event_id=event_id, deployment_id=model)
        
        return AzureOpenAIChatProvider(event_id=event_id, deployment_id=model)
\end{lstlisting}
\begin{lstlisting}[language=diff, label={lst:editbench_ours_step_1_diff}, caption=\sys extracts a helper function in Step 1 as a semantics-preserving refactor.]
--- sections/case_study/editbench/task_42_input.py	2026-02-24 14:58:47
+++ sections/case_study/editbench/task_42_ours_step1.py	2026-02-24 14:53:25
@@ -10,6 +10,7 @@
 
 # pylint: disable=W1203, C0415 [Use %s formatting in logging function, import-outside-toplevel]
 
+import asyncio
 
 class LLMArena(BaseModel):
     """
@@ -37,15 +38,16 @@
             provider_for_model = LLMArena._get_provider_for_model(model, **kwargs)
             providers.append(provider_for_model)
 
-        
-        for provider in providers:
+        async def generate_with_catch(provider, prompt):
             try:
-                response = await provider.generate_chat_response(prompt)
-                responses.append(response)
+                return await provider.generate_chat_response(prompt)
             except Exception as e:
-                logging.error(f"Error generating response from {provider}: {e}")
-                responses.append(f"Error generating response from {provider}: {e}")
+                msg = f"Error generating response from {provider}: {e}"
+                logging.error(msg)
+                return msg
 
+        responses = [await generate_with_catch(provider, prompt) for provider in providers]
+
         return responses
     
 
\end{lstlisting}
\begin{lstlisting}[language=diff, label={lst:editbench_ours_step_2_diff}, caption=\sys parallelizes the helper calls using \texttt{asyncio.gather} in Step 2.]
--- sections/case_study/editbench/task_42_ours_step1.py	2026-02-24 14:53:25
+++ sections/case_study/editbench/task_42_ours_step2.py	2026-02-24 14:52:59
@@ -46,8 +46,8 @@
                 logging.error(msg)
                 return msg
 
-        responses = [await generate_with_catch(provider, prompt) for provider in providers]
-
+        tasks = [generate_with_catch(provider) for provider in providers]
+        responses = await asyncio.gather(*tasks)
         return responses
     
 
\end{lstlisting}
\begin{lstlisting}[language=diff, label={lst:editbench_direct_step_1_diff}, caption=The direct transformation parallelizes using \texttt{asyncio.as\_completed} without preserving output order and misses the necessary import.]
--- sections/case_study/editbench/task_42_input.py	2026-02-24 14:58:47
+++ sections/case_study/editbench/task_42_direct_step1.py	2026-02-24 14:57:36
@@ -37,15 +37,15 @@
             provider_for_model = LLMArena._get_provider_for_model(model, **kwargs)
             providers.append(provider_for_model)
 
-        
-        for provider in providers:
+        tasks = [provider.generate_chat_response(prompt) for provider in providers]
+        for task in asyncio.as_completed(tasks):
             try:
-                response = await provider.generate_chat_response(prompt)
+                response = await task
                 responses.append(response)
             except Exception as e:
+                provider = providers[tasks.index(task)]
                 logging.error(f"Error generating response from {provider}: {e}")
                 responses.append(f"Error generating response from {provider}: {e}")
-
         return responses
     
 
\end{lstlisting}
\begin{lstlisting}[language=diff, label={lst:editbench_direct_step_2_diff}, caption=The direct transformation resolves the missing import.]
--- sections/case_study/editbench/task_42_direct_step1.py	2026-02-24 14:57:36
+++ sections/case_study/editbench/task_42_direct_step2.py	2026-02-24 14:59:23
@@ -9,8 +9,8 @@
 from carvana_enzo_worker.providers.azure_gpt_provider import AzureOpenAIChatProvider
 
 # pylint: disable=W1203, C0415 [Use %s formatting in logging function, import-outside-toplevel]
+import asyncio
 
-
 class LLMArena(BaseModel):
     """
     A tool to generate chats using multiple LLM's for a given prompt
\end{lstlisting}

\section{KernelBench Case Study}
\label{app:kernelbench_case_study}
We compare \sys results with the top solutions from the KernelBench leaderboard\footnote{https://scalingintelligence.stanford.edu/KernelBenchLeaderboard}.
To eliminate hardware variance, we run all solutions on a local L40S GPU and compute speedup relative to the baseline measured on the same machine.
\subsection{Matmul with Diagonal Matrices}
The task is to optimize a model that performs matrix multiplication of a diagonal matrix with another matrix.
\sys achieves a 13.2$\times$ speedup, while the leaderboard baseline (Claude-3.5 Sonnet) achieves a 12.2$\times$ speedup.
\paragraph{Memory coalescing.} 
\sys uses a 1D linear index that maps threads directly to adjacent memory addresses. 
This allows the GPU to "coalesce" multiple memory requests into a single transaction, maximizing throughput.
Claude’s 2D block mapping (16$\times$16) introduces extra coordinate arithmetic and can lead to less efficient memory access if matrix dimensions are not perfectly aligned.
\paragraph{Allocation overhead.} 
\sys uses \texttt{torch::empty\_like}, which allocates memory without clearing it. 
Claude uses \texttt{torch::zeros}, which performs a redundant memory pass to fill the tensor with zeros before the kernel runs. 
Since the kernel overwrites every value anyway, this extra pass in Claude-generated output wastes memory bandwidth.

\paragraph{Kernel simplicity.} 
The 1D grid in \sys simplifies the index calculation ($idx = i \cdot M + j$), reducing the number of integer instructions per thread compared to the 2D logic ($row$ and $col$ calculations) used in the Claude-generated output.

\begin{lstlisting}[language=Python, label={lst:kernelbench_leaderboard}, caption=Claude-3.5 Sonnet for ``Matmul with Diagonal Matrices''.]
import torch
import torch.nn as nn
from torch.utils.cpp_extension import load_inline

diag_matmul_cuda_source = """
#include <torch/extension.h>
#include <cuda_runtime.h>

__global__ void diag_matmul_kernel(
    const float* diag,
    const float* mat,
    float* out,
    const int N,
    const int M) {
    
    const int row = blockIdx.y * blockDim.y + threadIdx.y;
    const int col = blockIdx.x * blockDim.x + threadIdx.x;
    
    if (row < N && col < M) {
        out[row * M + col] = diag[row] * mat[row * M + col];
    }
}

torch::Tensor diag_matmul_cuda(torch::Tensor diag, torch::Tensor mat) {
    const int N = diag.size(0);
    const int M = mat.size(1);
    
    auto out = torch::zeros({N, M}, mat.options());
    
    const dim3 threads(16, 16);
    const dim3 blocks((M + threads.x - 1) / threads.x,
                     (N + threads.y - 1) / threads.y);
                     
    diag_matmul_kernel<<<blocks, threads>>>(
        diag.data_ptr<float>(),
        mat.data_ptr<float>(),
        out.data_ptr<float>(),
        N, M);
        
    return out;
}
"""

diag_matmul_cpp_source = """
torch::Tensor diag_matmul_cuda(torch::Tensor diag, torch::Tensor mat);
"""

diag_matmul = load_inline(
    name='diag_matmul',
    cpp_sources=diag_matmul_cpp_source,
    cuda_sources=diag_matmul_cuda_source,
    functions=['diag_matmul_cuda'],
    verbose=True
)

class ModelNew(nn.Module):
    def __init__(self):
        super(ModelNew, self).__init__()
        self.diag_matmul = diag_matmul
        
    def forward(self, A, B):
        return self.diag_matmul.diag_matmul_cuda(A.cuda(), B.cuda())
\end{lstlisting}
\begin{lstlisting}[language=Python, label={lst:kernelbench_ours}, caption=\sys for ``Matmul with Diagonal Matrices''.]
import torch
import torch.nn as nn
from torch.utils.cpp_extension import load_inline
 

diag_mult_source = """
 #include <torch/extension.h>
 #include <cuda_runtime.h>
 

 __global__ void diag_mult_kernel(const float* A, const float* B, float* out, int N, int M) {
  int idx = blockIdx.x * blockDim.x + threadIdx.x;
  if (idx < N * M) {
  int i = idx / M;
  int j = idx % M;
  out[idx] = A[i] * B[idx]; // B is stored in row-major order (B[i*M + j] is B[i][j])
  }
 }
 

 torch::Tensor diag_mult_cuda(torch::Tensor A, torch::Tensor B) {
  // Validate input dimensions
  TORCH_CHECK(A.dim() == 1 && B.dim() == 2, "A must be 1D and B must be 2D");
  TORCH_CHECK(A.size(0) == B.size(0), "A's length must match B's row count");
 

  int N = A.size(0);
  int M = B.size(1);
 

  auto out = torch::empty_like(B);
  const int threads_per_block = 256;
  int blocks = (N * M + threads_per_block - 1) / threads_per_block;
 

  diag_mult_kernel<<<blocks, threads_per_block>>>(
  A.data_ptr<float>(),
  B.data_ptr<float>(),
  out.data_ptr<float>(),
  N, M
  );
  return out;
 }
"""
 

diag_mult_cpp_source = (
  "torch::Tensor diag_mult_cuda(torch::Tensor A, torch::Tensor B);"
)
 

 # Compile the inline CUDA code
diag_mult = load_inline(
  name="diag_mult",
  cpp_sources=diag_mult_cpp_source,
  cuda_sources=diag_mult_source,
  functions=["diag_mult_cuda"],
  verbose=True,
)
 

class ModelNew(nn.Module):
  def __init__(self):
    super().__init__()
    self.diag_mult = diag_mult
 

  def forward(self, A, B):
    return self.diag_mult.diag_mult_cuda(A, B)
\end{lstlisting}

\subsection{Conv2D Subtract Subtract Mish}
The task is to optimize a model that performs a convolution, subtracts two scalar values element-wise, and applies Mish activation.
\sys achieves 1.76$\times$ speedup, while the leaderboard baseline (DeepSeek Coder) achieves 0.78$\times$ slowdown.

\paragraph{Kernel fusion.}
DeepSeek Coder uses three separate CUDA kernels (elementwise subtract $\times$2, Mish), each requiring a kernel launch and writing intermediate results to global memory.
\sys fuses the two subtracts and Mish into a single kernel.
This reduces kernel launch overhead and avoids intermediate global-memory round-trips; values remain in registers across the entire computation.

\paragraph{Algebraic simplification.}
DeepSeek Coder chains two subtract kernels: $x \to x - v_1 \to (x - v_1) - v_2$.
\sys recognizes the semantic equivalence $(x - v_1) - v_2 = x - (v_1 + v_2)$ and applies a single offset, combining both subtractions into one operation inside the fused kernel.

\paragraph{Allocation overhead.}
Both elementwise-subtract and Mish kernels in DeepSeek Coder use \texttt{torch::zeros\_like}, which zero-initializes memory before the kernel overwrites it.
\sys uses \texttt{torch::empty\_like}, which allocates without initialization, eliminating redundant memory writes since every element is overwritten by the kernel.

\begin{lstlisting}[language=Python, label={lst:kernelbench_conv2d_deepseek}, caption=DeepSeek Coder for ``Conv2D Subtract Subtract Mish''.]
import torch
import torch.nn as nn
import torch.nn.functional as F
from torch.utils.cpp_extension import load_inline

# Define the custom CUDA kernel for element-wise subtraction
elementwise_subtract_source = """
#include <torch/extension.h>
#include <cuda_runtime.h>

__global__ void elementwise_subtract_kernel(const float* x, float* out, float value, int size) {
    int idx = blockIdx.x * blockDim.x + threadIdx.x;
    if (idx < size) {
        out[idx] = x[idx] - value;
    }
}

torch::Tensor elementwise_subtract_cuda(torch::Tensor x, float value) {
    auto size = x.numel();
    auto out = torch::zeros_like(x);

    const int block_size = 256;
    const int num_blocks = (size + block_size - 1) / block_size;

    elementwise_subtract_kernel<<<num_blocks, block_size>>>(x.data_ptr<float>(), out.data_ptr<float>(), value, size);

    return out;
}
"""

elementwise_subtract_cpp_source = "torch::Tensor elementwise_subtract_cuda(torch::Tensor x, float value);"

# Compile the inline CUDA code for element-wise subtraction
elementwise_subtract = load_inline(
    name='elementwise_subtract',
    cpp_sources=elementwise_subtract_cpp_source,
    cuda_sources=elementwise_subtract_source,
    functions=['elementwise_subtract_cuda'],
    verbose=True,
    extra_cflags=[''],
    extra_ldflags=['']
)

# Define the custom CUDA kernel for Mish activation
mish_source = """
#include <torch/extension.h>
#include <cuda_runtime.h>
#include <math.h>

__global__ void mish_kernel(const float* x, float* out, int size) {
    int idx = blockIdx.x * blockDim.x + threadIdx.x;
    if (idx < size) {
        float exp_x = expf(x[idx]);
        float softplus = logf(1.0f + exp_x);
        out[idx] = x[idx] * tanhf(softplus);
    }
}

torch::Tensor mish_cuda(torch::Tensor x) {
    auto size = x.numel();
    auto out = torch::zeros_like(x);

    const int block_size = 256;
    const int num_blocks = (size + block_size - 1) / block_size;

    mish_kernel<<<num_blocks, block_size>>>(x.data_ptr<float>(), out.data_ptr<float>(), size);

    return out;
}
"""

mish_cpp_source = "torch::Tensor mish_cuda(torch::Tensor x);"

# Compile the inline CUDA code for Mish activation
mish = load_inline(
    name='mish',
    cpp_sources=mish_cpp_source,
    cuda_sources=mish_source,
    functions=['mish_cuda'],
    verbose=True,
    extra_cflags=[''],
    extra_ldflags=['']
)

class ModelNew(nn.Module):
    """
    Optimized model that performs a convolution, subtracts two values, applies Mish activation using custom CUDA kernels.
    """
    def __init__(self, in_channels, out_channels, kernel_size, subtract_value_1, subtract_value_2):
        super(ModelNew, self).__init__()
        self.conv = nn.Conv2d(in_channels, out_channels, kernel_size)
        self.subtract_value_1 = subtract_value_1
        self.subtract_value_2 = subtract_value_2
        self.elementwise_subtract = elementwise_subtract
        self.mish = mish

    def forward(self, x):
        x = self.conv(x)
        x = self.elementwise_subtract.elementwise_subtract_cuda(x, self.subtract_value_1)
        x = self.elementwise_subtract.elementwise_subtract_cuda(x, self.subtract_value_2)
        x = self.mish.mish_cuda(x)
        return x
\end{lstlisting}
\begin{lstlisting}[language=Python, label={lst:kernelbench_conv2d_ours}, caption=\sys for ``Conv2D Subtract Subtract Mish''.]
import torch
import torch.nn as nn
from torch.utils.cpp_extension import load_inline

class ModelNew(nn.Module):
    def __init__(self, in_channels, out_channels, kernel_size, subtract_value_1, subtract_value_2):
        super(ModelNew, self).__init__()
        self.conv = nn.Conv2d(in_channels, out_channels, kernel_size)
        self.offset = subtract_value_1 + subtract_value_2  # Combined subtraction offset
        
        # Define and load the fused subtract + Mish CUDA kernel
        fused_kernel_source = """
        #include <torch/extension.h>
        #include <cuda_runtime.h>
        #include <math_constants.h>

        __global__ void fused_subtract_mish_kernel(
            const float* input, float offset, float* output, int size
        ) {
            int idx = blockIdx.x * blockDim.x + threadIdx.x;
            if (idx < size) {
                float temp = input[idx] - offset;
                float exp_temp = expf(temp);
                float softplus_val = log1pf(exp_temp);
                float tanh_sp = tanhf(softplus_val);
                output[idx] = temp * tanh_sp;
            }
        }

        torch::Tensor fused_subtract_mish_cuda(torch::Tensor input, float offset) {
            auto size = input.numel();
            auto output = torch::empty_like(input);
            
            const int threads_per_block = 256;
            const int blocks_per_grid = (size + threads_per_block - 1) / threads_per_block;

            fused_subtract_mish_kernel<<<blocks_per_grid, threads_per_block>>>(
                input.data_ptr<float>(), offset, output.data_ptr<float>(), size
            );
            return output;
        }
        """

        fused_kernel_cpp = (
            "torch::Tensor fused_subtract_mish_cuda(torch::Tensor input, float offset);"
        )

        self.fused_kernel = load_inline(
            name="fused_subtract_mish",
            cpp_sources=fused_kernel_cpp,
            cuda_sources=fused_kernel_source,
            functions=["fused_subtract_mish_cuda"],
            verbose=False
        )

    def forward(self, x):
        conv_output = self.conv(x)
        # Apply fused subtract and Mish in a single CUDA kernel
        return self.fused_kernel.fused_subtract_mish_cuda(conv_output, self.offset)
\end{lstlisting}

\section{Repo-Level Case Study}
\label{app:sweperf_case_study}
We use SWE-Perf~\cite{he2025swe} for our repo-level case study.
Specifically, we present the task \texttt{astropy\_\_astropy-13734}, which involves optimizing the fixed-width ASCII table parsing code in the Astropy library.

\sys achieves 0.20 statistical performance gain~\cite{he2025swe} during the semantics-preserving improvement and 0.32 after the performance-oriented transformation, while DeepSeek-V3-Chat achieves 0.15 after two iterations of performance-oriented transformation.

Both \sys and DeepSeek-V3-Chat address a common bottleneck by caching processed lines instead of calling \texttt{self.process\_lines(lines)} twice.
The key difference lies in column width computation: DeepSeek-V3-Chat makes repeated passes over the rows, while \sys uses a single pass with sequential row-wise access, resulting in better cache locality.

\begin{lstlisting}[language=diff, label={lst:sweperf-deepseek-v3}, caption=DeepSeek-V3 on SWE-Perf.]
diff --git a/astropy/io/ascii/fixedwidth.py b/astropy/io/ascii/fixedwidth.py
index 9a2d322..d5e5233 100644
--- a/astropy/io/ascii/fixedwidth.py
+++ b/astropy/io/ascii/fixedwidth.py
@@ -34,11 +34,14 @@ class FixedWidthSplitter(core.BaseSplitter):
     delimiter = '|'
 
     def __call__(self, lines):
-        for line in lines:
-            vals = [line[x.start:x.end] for x in self.cols]
-            if self.process_val:
-                yield [self.process_val(x) for x in vals]
-            else:
+        process_val = self.process_val
+        if process_val:
+            for line in lines:
+                vals = [line[x.start:x.end] for x in self.cols]
+                yield [process_val(x) for x in vals]
+        else:
+            for line in lines:
+                vals = [line[x.start:x.end] for x in self.cols]
                 yield vals
 
     def join(self, vals, widths):
@@ -91,10 +94,12 @@ class FixedWidthHeader(basic.BasicHeader):
             List of table lines
 
         """
+        # Process lines once and cache the result to avoid reprocessing.
+        processed_lines = list(self.process_lines(lines))
 
         # See "else" clause below for explanation of start_line and position_line
-        start_line = core._get_line_index(self.start_line, self.process_lines(lines))
-        position_line = core._get_line_index(self.position_line, self.process_lines(lines))
+        start_line = core._get_line_index(self.start_line, processed_lines)
+        position_line = core._get_line_index(self.position_line, processed_lines)
 
         # If start_line is none then there is no header line.  Column positions are
         # determined from first data line and column names are either supplied by user
@@ -130,7 +135,7 @@ class FixedWidthHeader(basic.BasicHeader):
                 # column names.  Note that get_fixedwidth_params returns Python
                 # slice col_ends but expects inclusive col_ends on input (for
                 # more intuitive user interface).
-                line = self.get_line(lines, position_line)
+                line = self.get_line(processed_lines, position_line)
                 if len(set(line) - {self.splitter.delimiter, ' '}) != 1:
                     raise InconsistentTableError(
                         'Position line should only contain delimiters and '
@@ -149,7 +154,7 @@ class FixedWidthHeader(basic.BasicHeader):
                 self.col_ends = [x - 1 if x is not None else None for x in col_ends]
 
             # Get the header column names and column positions
-            line = self.get_line(lines, start_line)
+            line = self.get_line(processed_lines, start_line)
             vals, starts, ends = self.get_fixedwidth_params(line)
 
             self.names = vals
@@ -238,17 +243,24 @@ class FixedWidthData(basic.BasicData):
     """ Splitter class for splitting data lines into columns """
 
     def write(self, lines):
-        vals_list = []
+        # Collect all values and compute column widths in one pass.
         col_str_iters = self.str_vals()
+        vals_list = []
+        # Initialize widths with zeros or with header name lengths if present.
+        if self.header.start_line is not None:
+            widths = [len(col.info.name) for col in self.cols]
+        else:
+            widths = [0] * len(self.cols)
+
         for vals in zip(*col_str_iters):
             vals_list.append(vals)
+            for i, val in enumerate(vals):
+                if len(val) > widths[i]:
+                    widths[i] = len(val)
 
+        # Set column widths.
         for i, col in enumerate(self.cols):
-            col.width = max(len(vals[i]) for vals in vals_list)
-            if self.header.start_line is not None:
-                col.width = max(col.width, len(col.info.name))
-
-        widths = [col.width for col in self.cols]
+            col.width = widths[i]
 
         if self.header.start_line is not None:
             lines.append(self.splitter.join([col.info.name for col in self.cols],
\end{lstlisting}

\begin{lstlisting}[language=diff, label={lst:sweperf-ours-sem}, caption=\sys performs the semantics-preserving transformation on SWE-Perf.]
diff --git a/astropy/io/ascii/fixedwidth.py b/astropy/io/ascii/fixedwidth.py
index 9a2d322..aa8c29f 100644
--- a/astropy/io/ascii/fixedwidth.py
+++ b/astropy/io/ascii/fixedwidth.py
@@ -51,7 +51,7 @@ class FixedWidthHeader(basic.BasicHeader):
 
         """
 
-        # See "else" clause below for explanation of start_line and position_line
-        start_line = core._get_line_index(self.start_line, self.process_lines(lines))
-        position_line = core._get_line_index(self.position_line, self.process_lines(lines))
-
+        # Preprocess lines once to avoid repeated processing
+        processed_lines = list(self.process_lines(lines))
+        start_line = core._get_line_index(self.start_line, processed_lines)
+        position_line = core._get_line_index(self.position_line, processed_lines)
         # If start_line is none then there is no header line.  Column positions are
         # determined from first data line and column names are either supplied by user
         # or auto-generated.
@@ -243,10 +243,22 @@ class FixedWidthData(basic.BasicData):
         for vals in zip(*col_str_iters):
             vals_list.append(vals)
 
+        # Compute column widths
+        lengths = []
+        for vals in vals_list:
+            lengths_row = [len(v) for v in vals]
+            lengths.append(lengths_row)
+
+        max_data = [max(col) for col in zip(*lengths)]
+
+        if self.header.start_line is not None:
+            name_lengths = [len(col.info.name) for col in self.cols]
+            max_lengths = [max(d, n) for d, n in zip(max_data, name_lengths)]
+        else:
+            max_lengths = max_data
+
         for i, col in enumerate(self.cols):
-            col.width = max(len(vals[i]) for vals in vals_list)
-            if self.header.start_line is not None:
-                col.width = max(col.width, len(col.info.name))
+            col.width = max_lengths[i]
 
         widths = [col.width for col in self.cols]
 
\end{lstlisting}

\begin{lstlisting}[language=diff, label={lst:sweperf-ours-perf}, caption=\sys performs optimization on SWE-Perf.]
diff --git a/astropy/io/ascii/fixedwidth.py b/astropy/io/ascii/fixedwidth.py
index 9a2d322..bb47725 100644
--- a/astropy/io/ascii/fixedwidth.py
+++ b/astropy/io/ascii/fixedwidth.py
@@ -92,10 +92,10 @@ class FixedWidthHeader(basic.BasicHeader):
 
         """
 
-        # See "else" clause below for explanation of start_line and position_line
-        start_line = core._get_line_index(self.start_line, self.process_lines(lines))
-        position_line = core._get_line_index(self.position_line, self.process_lines(lines))
-
+        # Preprocess lines once to avoid repeated processing
+        processed_lines = list(self.process_lines(lines))
+        start_line = core._get_line_index(self.start_line, processed_lines)
+        position_line = core._get_line_index(self.position_line, processed_lines)
         # If start_line is none then there is no header line.  Column positions are
         # determined from first data line and column names are either supplied by user
         # or auto-generated.
@@ -243,11 +243,20 @@ class FixedWidthData(basic.BasicData):
         for vals in zip(*col_str_iters):
             vals_list.append(vals)
 
+        # Compute column widths in a single pass for efficiency
+        num_cols = len(self.cols)
+        max_lengths = [0] * num_cols
+        for vals in vals_list:
+            for i, val in enumerate(vals):
+                val_len = len(val)
+                if val_len > max_lengths[i]:
+                    max_lengths[i] = val_len
+
         for i, col in enumerate(self.cols):
-            col.width = max(len(vals[i]) for vals in vals_list)
+            width = max_lengths[i]
             if self.header.start_line is not None:
-                col.width = max(col.width, len(col.info.name))
-
+                width = max(width, len(col.info.name))
+            col.width = width
         widths = [col.width for col in self.cols]
 
         if self.header.start_line is not None:
\end{lstlisting}


\end{document}